# Quantifying out-of-distribution data severity with Wasserstein distances: benchmarking large language models for alloy property prediction

YuJie Liu[1], Dezhen Xue[1], Jun Sun[1], Xiangdong Ding[1], Zhibin Gao[1†] and Turab Lookman[1,2*]

[1]State Key Laboratory for Mechanical Behavior of Materials, State Key Laboratory of Porous Metal Materials, School of Materials Science and Engineering, Xi'an Jiaotong University, Xi'an 710049, China
[2]AiMaterials Research, Santa Fe, New Mexico, USA 87501
†E-mail: zhibin.gao@xjtu.edu.cn
*E-mail: turablookman@gmail.com

## Abstract

General-purpose Large Language Models (LLMs) increasingly rival domain-specific pipelines on scientific prediction, but their performance under distribution shift is untested. Across six alloy and crystal-structure datasets, we benchmark LLMs against traditional, BERT-based, TabPFN, and text-regression baselines under random cross-validation and six out-of-distribution (OOD) protocols shifting input and target distributions, with severity quantified via a standardized Wasserstein distance. We isolate the LLM's advantage over k-nearest-neighbor (k-NN) retrieval: it beats k-NN on all eight tasks at a fixed 5% reference budget, on seven of eight when independently optimized, and in 72.5% of 120 OOD task–protocol configurations against an optimized baseline. No model family is uniformly best; the LLM is often strongest under severe shift, and error increases with target-space but not representation-space severity. Self-reported LLM confidence tracks both shift and error, more consistently than a density-only control from the same retrieved samples, and a matched-cohort check finds no substantial evidence of memorization.

**Introduction**

The rapid advance of machine learning in materials science has traditionally emphasized domain-specific models pretrained on vast corpora of specialized literature, such as journal articles, reports, and databases[1–7]. This approach stems from the prevailing view that extensive, targeted data is essential for accurate predictions, with techniques such as fine-tuning[8–10] and Retrieval-Augmented Generation (RAG)[11] further refining these models for specific tasks[12–16]. Pretrained models tailored to materials classes—such as SteelBERT[4] for steels, MatSciBERT[17] for broader materials science, and SciBERT[18] for scientific text—have shown promise to varying degrees in encoding domain knowledge, enabling downstream tasks such as synthesis protocol extraction[1,19–25] and property prediction[26–28].

However, developments in natural language processing have shifted this landscape. Optimization strategies, such as Reinforcement Learning (RL) during fine-tuning or pretraining, have endowed general-purpose Large Language Models (LLMs) with superior reasoning capabilities[29–32]. Trained on diverse, internet-scale datasets encompassing trillions of tokens, Gemini and DeepSeek can perform complex inference without explicit domain adaptation[19,33]. Furthermore, the direct application of LLMs to materials property prediction has recently gained significant traction. It has been shown that a language model (LLM-Prop[34]) based on a transformer T5 architecture and trained on human-readable textual descriptions of crystalline structures can capture key relationships between crystal structure and material properties, such as band gap and formation energy. LLM-Prop further showed that language-based structural descriptions can be highly predictive, outperforming strong graph neural network baselines (GNNs), as well as a materials specific model, MatBERT, on several Materials Project tasks. However, LLM-Prop relies on task-specific fine-tuning of a T5 encoder, leaving open whether general-purpose LLMs can achieve similar predictive power without retraining. Nevertheless, LLM4Mat-Bench[35] addressed this more broadly, testing the chat-based LLMs Llama, Mistral and Gemma on regression and classification tasks across multiple crystal datasets, properties and input modalities, such as compositions, CIFs and textual descriptions. In general, the fine-tuned domain models usually outperform off-the-shelf chat LLMs in numerical prediction, under zero-shot or few-shot settings[2,4,35].

Recently, the transformer TabPFN[36] trained on over 130 million synthetically generated tabular data has been introduced to allow rapid prediction of objectives. The prior fitted model learns through in context learning (ICL) on unmasked samples and can be applied to an arbitrary unseen real-world dataset in one pass. It shows considerable promise, outperforming baseline models, e.g. tree based methods, linear models, SVMs, MLPs, an ensemble model CatBoost[37] and AutoGluon[38], a stacked ensemble of base line machine learning models. It is also robust enough to handle datasets of 10,000 samples, 500 features and 10 classes, and these limits have been extended in recent updates. We have employed TabPFN as a benchmark model to compare LLMs with.

For structural materials in general, and multicomponent alloys in particular, few datasets based on mere calculations or crystal structure exist that faithfully capture experimental properties. The data are painstakingly acquired in the laboratory and are therefore relatively not extensive, typically of the order of $10^2$-$10^3$, as the properties depend not only on compositions but also on processing history, microstructure, and testing conditions[39]. Although pretrained materials models have demonstrated that language models can encode composition–process–property relations, their robustness across alloy families remain insufficiently tested [4,5,40]. Therefore, there is a need to evaluate LLMs for materials property prediction against strong baselines such as GNNs, tree ensembles, and tabular foundation models under realistic datasets with explicit processing variables[41].

A key to unlocking general-purpose LLMs without expensive fine-tuning has been shown to lie in advanced prompt engineering and in-context learning (ICL). Providing LLMs with a small number of high-quality examples in few shot learning can significantly mitigate data scarcity, allowing models to identify patterns without extensive retraining[42,43]. For example, Polak and Morgan utilized conversational redundancy to extract accurate data from text without fine-tuning[23], while the MatAgent framework demonstrated that LLMs could orchestrate external tools for property retrieval[44]. Wang et al.[45] evaluated the robustness of LLMs in predicting steel yield strengths, to reveal that while retrieving "nearest neighbor" examples improves accuracy, pretrained LLMs often suffer from "mode collapse"—a tendency to default to high-frequency training values when the input distribution shifts. Critically, their approach relies on the LLM to perform the regression task purely from the provided context, which exposes the model's inherent weakness in precise arithmetic and extrapolation. This raises a key question: can these general LLMs rival or surpass models pretrained for materials property prediction, particularly for metallic alloys where data scarcity and compositional and processing complexity pose challenges?

In this work, we address this issue by systematically comparing general-purpose LLMs (Gemini, DeepSeek, GPT) against domain-specific pretrained (SteelBERT, MatSciBERT, SciBERT) and benchmark models LLM-Prop from LLM4MatBench, MODNet[46], TPOT-Mat[47] and GPTChem[48] from MatBench[41], and TabPFN across datasets for several alloys including Titanium (Ti), High-entropy alloy (HEA), Aluminum (Al) and Steels, together with a Matbench steels dataset and a crystal-structure dataset, for a total of six datasets spanning composition- and structure-based property prediction. The sizes of the experimental datasets range from over 200 samples with composition and processing for the ultimate tensile strength (UTS) and elongation for Ti to over 600 for Steels. Our benchmark of crystal structure properties is also limited to the ~ 600 data samples.

The trained patterns learned by a model are from the training or "in-distribution" data, and data that is out of this distribution (out of distribution, OOD) will not be expected to obey the model and will often give rise to unexpected results. Recently, the performance of models on various OODs has been examined using by and large traditional machine learning models[49–52]. Omee et al.[52] conducted a benchmark study of graph neural network (GNN) based algorithms on three datasets from Matbench pertaining to inorganic crystal structures using five sets of OOD test data. Their general conclusion was that no algorithm triumphed in all situations and on average their performance was worse on the OOD data compared to the baselines MatBench. Similarly, Li et al.[53] examined the performance of a range of traditional ML models, as well as LLM-Prop on OOD data generated using chemical and symmetry identity splitting. They use databases from Materials project, OQMD and JARVIS, and conclude that many of the model predictions on test data were robust and not a challenge to the models. Rather, they appeared to be overestimates to model generalization. An issue with these studies is that they primarily evaluated model predictions, and changes to the geometry of the distributions of the training and test data were inferred by visualizing the two dimensional projections of the materials representations. There was no quantitative measure to determine the extent to which the spatial geometry, as well as the properties, of the test and training distributions differed.

To rigorously quantify the novelty of any test sample or distribution and its proximity to the training manifold, we utilize the Wasserstein distance, a metric rooted in the mathematical framework of Optimal Transport (OT) theory[54]. Obtained by solving the Monge-Kantorovich optimization problem, the Wasserstein distance (commonly also called the Earth Mover's Distance) represents the minimum "work"—defined as the product of mass and ground distance—required to transform one probability distribution into another[55]. While computationally intensive historically, modern algorithmic regularizations have made it highly scalable for high-dimensional features. Unlike the information-theoretic measure Kullback-Leibler (KL) divergence, which relies on pointwise density overlaps and can become uninformative or undefined when two distributions have disjoint supports, the Wasserstein distance incorporates the underlying geometry of the feature space[56]. This allows it to remain a smooth and informative metric even when comparing entirely distinct material families. Moreover, Redko et al.[57] proved in their landmark paper on domain-adaptation theory that a model's generalization error on a target dataset is bounded by the Wasserstein distance between the target and the source dataset. Thus, in the context of materials discovery, this geometric approach enables a robust characterization of OOD samples by quantifying the literal distance a model's reasoning must "bridge" to map a novel alloy composition back to the known training data. The distance has been previously used to also quantify similarity in the context of crystal structures and composition profiles[58,59].

Here, we evaluate model performance for the alloy datasets under OOD settings using two types of test protocols: OOD test distributions using the five structural splitting methods adopted in Ref. [52], one of which employs cluster analysis studied in[51], and extrapolation ability by selecting the top 20% of samples with the highest target values as the test set and the remaining 80% for training. This extrapolation split is designed to evaluate whether models can predict high-property regions that are absent from the training distribution.

Because LLM-based prediction and k-NN retrieval both draw on the same pool of similar training examples, we directly test what the LLM adds beyond retrieval alone. At a fixed 5% reference budget, the LLM outperforms the best k-NN baseline on all eight alloy tasks, by 23.78–34.35% in MAE; when the k-NN baseline is instead independently optimized over reference size, representation and weighting, and the LLM is given exactly the same references, it still wins on seven of the eight tasks. Under every out-of-distribution protocol used in this study, against a k-NN baseline tuned directly on the evaluation data — an optimistic comparison that favours k-NN — the LLM remains ahead in 61.67% of 120 task-protocol configurations, most consistently under compositional shift (leave-one-cluster-out, LOCO) and least consistently for isolated low-density test points.

The manuscript is organized as follows: In the LLM, we employ RAG with similarity or "Querying for Maximally Informative Samples" (via all-MiniLM-L6-v2 embeddings) to prioritize data quality over quantity[45]. We demonstrate that 5% of highly relevant alloy reference samples can match or exceed the performance of full datasets. We also perform a prompt ablation study to determine whether the performance is influenced by the prompt order/design or originates from the LLM itself. Moreover, choosing samples "similar" to the target reduces data requirements significantly as merely 5% of the training data is now necessary to make adequate predictions. We also study how prediction accuracy can be improved using a "Guided mode," in which a prior estimate from another model is applied alongside the reference samples in the prompt. For LLM versus k-NN, we directly test what the LLM adds beyond simple k-NN retrieval, comparing matched-reference and independently-tuned k-NN baselines both in-distribution and under every out-of-distribution protocol used in this study. Here, we focus on OOD generalization and introduce the Wasserstein distance as an appropriate metric to characterize the departure away from the training data. Using this metric, we determine that extrapolation and clustering, in general, provide the most severe tests to a model with GPT-5.5 strongest under "extreme" OOD shifts. However, the "best" model is determined not only by the alloy dataset and target property, but also by the specific type of OOD shift used to define the test set. Finally, as the OOD severity increases, the fraction of high-confidence LLM reported predictions is reduced with increasing statistical prediction errors, while lower relative-error outcomes are associated with high-confidence predictions. We further find that this confidence–error relationship is not simply a byproduct of retrieval proximity: a density-only control built from the same retrieved samples tracks error less consistently than the LLM's self-reported confidence.

To be explicit about scope, this paper is a benchmark of general-purpose and domain-specific language models against established tabular and materials-informatics baselines for alloy property prediction, together with a quantitative measure of out-of-distribution severity and its relationship to prediction error. It does not propose a new prediction framework: the retrieval-and-refinement workflow introduced above is evaluated as one configuration among several candidate methods, not advanced as a contribution in its own right.

Correspondingly, the claim that the full analysis protocol — the multi-step prompting scaffold that structures the LLM's reasoning (baseline anchoring, plausibility assessment, interpolative correction), distinct from the out-of-distribution splitting protocols introduced below — is required for accurate prediction should be read as conditional rather than general. The Matbench comparison shows this directly (Tables 5, 6): removing this prompting scaffold changes GPT-5.4's MAE only marginally (93.65 versus 95.64 MPa on the steels task), and a fuller ablation of the same scaffold across multiple model generations, reported in the Fig. S2, shows the same pattern more broadly — newer, more capable models depend on it far less than older ones, for which it still provides a measurable benefit. The analysis protocol remains a useful diagnostic tool and a common baseline for comparing models, but its necessity is a property of the specific model being evaluated, not a general requirement for LLM-based property prediction.

## Results

### Datasets overview

In this benchmark, we considered five alloy datasets and one crystal-structure dataset targeting exfoliation energy. The dataset size, input representation, and data source for each dataset are summarized in Table 1, while the complete list of input feature columns is provided in Table S1. These datasets cover diverse forms of materials representations. Specifically, the Steel dataset encodes processing conditions as sequential heat-treatment descriptions, while Matbench-Steel-YS relies solely on elemental compositions. For the JDFT2D dataset, crystal structures were transformed into textual descriptions using Robocrystallographer[60], which were then used as LLM inputs. The Aluminium, Titanium, and high-entropy alloy datasets provide richer structured representations, incorporating both compositional information and processing parameters.

**Table 1** Overview of datasets and prediction tasks.

| Label | Target | n | Inputs | Provenance |
|---|---|---|---|---|
| Aluminium | UTS | 785 | Composition (18 wt%) + 13 processing | Literature |
| Titanium | UTS,Elongation | 233 | Composition (9 wt%) + 6 processing | Literature |
| High-entropy alloys | UTS | 491 | Composition (14 at%) + 8 processing | Literature |
| Steels | UTS,Elongation | 677 | Composition (36 wt%) + text | Literature |
| Matbench-Steel-YS | Yield strength | 312 | Composition (14 at%) | Matbench (MatbenchSteelYS) |
| Matbench-JDFT2D | DFT exfoliation energy | 636 | Structure → text (Robocrystallographer) | Matbench (MatbenchJDFT2DExfoliation) |

### Outline of experiments

**Table 2** Summary of LLM configurations and experimental protocols used in this study. The table reports the LLM versions, Guided or Unguided, prompting protocols, reference samples, and number of repeated runs for each experiment. Reference settings vary according to the purpose underlying each analysis, including reference-ratio

studies, KNN comparisons, in-distribution evaluations, and out-of-distribution (OOD) assessments. Unless otherwise specified, all LLM experiments use the No protocol configuration, which does not prescribe an explicit step-by-step inference procedure. GPT-5.4, Gemini-2.5-Flash, Gemini-2.5-Pro and Gemini-3.0-Flash were deprecated in the course of this work.

| Location of analysis | LLM(s) and version | Reference samples | Repeats |
|---|---|---|---|
| **Assessing potential pretraining leakage with a matched recent cohort** | GPT-5.5 | Full training set | 3 |
| **In-distribution alloy benchmark** | GPT-5.4 | Full training set | 1 |
| **In-distribution alloy benchmark** | GPT-5.5 | Full training set | 3 |
| **Matbench comparison** | GPT-5.5 | Full training set | 3 |
| **Matbench comparison** | Gemini-3-Flash Preview; GPT-5.4 | Full training set | 1 |
| **Similarity-based retrieval and reference ratio** | Gemini-2.5-Pro; Gemini-2.5-Flash | 0.01–0.10 step 0.01; 0.2–1.0 step 0.1 | 1 |
| **Guided vs Unguided** | Gemini-2.5-Pro; Gemini-2.5-Flash | 0.01–0.10 step 0.01; 0.2–1.0 step 0.1 | 1 |
| **Influence of representation space on LLM and k-NN prediction performance** | GPT-5.5; Deepseek-V4.1-Flash | 5% training pool | 3 |
| **LLM versus k-NN at a fixed reference budget** | Gemini-2.5-Pro; Gemini-2.5-Flash | 5 % of training pool | 1 |
| **LLM versus k-NN at a fixed reference budget** | GPT-5.5 | 5 % of training pool | 3 |
| **Optimizing the reference-set size** | GPT-5.5 | Best-performing k of k-NN (test-set MAE) | 3 |
| **Benchmarking model robustness under OOD shifts** | GPT-5.5 | Best-performing k of k-NN (test-set MAE) | 3 |
| **Benchmarking model robustness under OOD shifts** | GPT-5.5 | Full training set | 3 |
| **Error against OOD severity: a quantitative relation** | GPT-5.5 | Full training set | 3 |
| **Optimal k-NN references and LLM-reported confidence** | Gemini-2.5-Flash; Gemini-3-Flash Preview | Full training set | 1 |
| **Optimal k-NN references and LLM-reported confidence** | GPT-5.5 | Best-performing k of k-NN (test-set MAE) | 3 |
| **Effect of OOD severity on LLM-reported confidence** | GPT-5.5 | Full training set | 3 |
| **Comparison of LLM confidence with retrieval density** | GPT-5.5 | Best-performing k of k-NN/Full training set | 3 |
| **Comparison of LLM confidence with retrieval density** | GPT-5.5 | Full training set | 3 |

The integration of Large Language Models (LLMs) into materials science marks a shift from traditional supervised learning toward semantic reasoning agents. This transition represents the next stage in the history of Active Learning[61], moving from the 'label-efficiency' goals of the 1990s toward the 'discovery-efficiency' goals of modern autonomous laboratories[62]. Unlike previous encoder-only architectures such as SciBERT[18], MatSciBERT[17] or SteelBERT[4], which rely on extensive pretraining or fine-tuning for specific tasks, generative models like Gemini, DeepSeek and GPT utilize in-context learning to perform few-shot reasoning. Recent literature demonstrates that when guided by structured prompts—specifically those employing Chain-of-Thought (CoT) reasoning—LLMs can effectively extract complex synthesis protocols from unstructured text[19–22,63,64]. However, general-purpose LLMs remain limited by "hallucinations". This issue has been explored within a RAG framework in which relevance-ranked contextual information is retrieved from an external materials dataset and incorporated into the LLM input[45]. The retrieved context consists of alloy samples with similar chemical compositions and processing descriptions, providing experimentally grounded reference examples for the model. Similarity-based retrieval using a sentence embedding model, a lightweight transformer trained via contrastive learning to generate dense semantic representations of short texts, can then be used in information retrieval tasks[45]. As illustrated in Figure 1, we first evaluate k-nearest neighbor (k-NN) regression across four material representations: the original numerical features, text embeddings corresponding to the LLM inputs, and two-dimensional projections obtained separately using principal component analysis (PCA) and uniform manifold approximation and projection (UMAP). For each representation, two neighbor-aggregation strategies—uniform averaging and similarity-weighted averaging—are considered. The number of nearest neighbors, $k$, is varied to identify the best-performing representation and the corresponding value of $k$. The selected representation is then held fixed for subsequent LLM-based predictions.

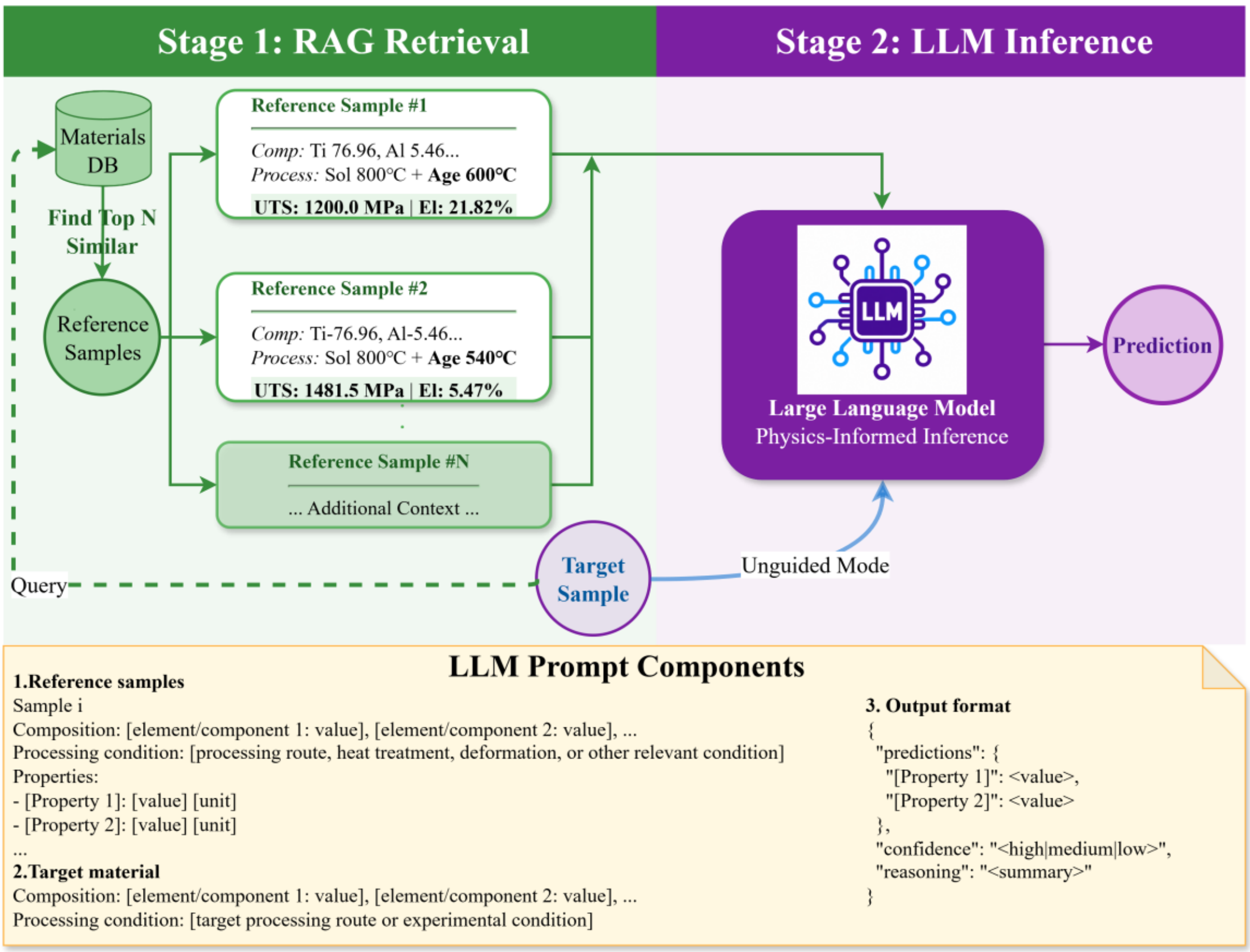

**Figure 1** Schematic overview of the three-stage hybrid workflow for mechanical property prediction. Stage 1, RAG: the Top-N most similar reference samples are retrieved from the database based on composition and processing information. Stage 2, LLM Inference: the LLM generates the final prediction by comparing the target sample with the retrieved reference samples. "Physics-informed inference" denotes the model's use of physically consistent composition–processing–property relationships drawn from the retrieved reference samples, not an independent mechanistic understanding of materials physics; a representative worked example is given in Supplementary Note.

For a given target sample, defined as an alloy composition–processing pair whose mechanical properties are to be predicted, we compute the cosine similarity $(\cos\theta)$ between its feature vector and those of all training samples within the selected representation space. The most similar samples are subsequently retrieved and incorporated into the LLM prompt as reference examples, allowing the predictions to be conditioned on experimental data relevant to the target composition and processing conditions. This configuration is referred to as the Unguided mode, in which no model-derived initial estimates are provided. By contrast, the Guided mode additionally supplies predictions from a pre-existing predictive model as initial estimates in the prompt, as illustrated in Fig. S1. The LLM draws on both these estimates and the retrieved reference samples to infer the target mechanical properties and generate final predictions.

As general-purpose LLMs develop stronger inference capabilities, explicit step-by-step prompting may become less essential for few-shot material-property prediction. To assess the contribution of these instructions, we compare the full prompt with the "No protocol" configuration. The full prompt specifies three inference steps (1&2&3, Figs. S2, 3), comprising reference-based baseline establishment, plausibility assessment, and interpolative correction, whereas "No protocol" omits these instructions. Both configurations retain the same task definition, retrieved reference samples, target material information, and output format. The complete ablation design, including the intermediate (1&2) configuration, is described in the Supplementary Prompt. As shown in Fig. S2, 3, explicit inference protocols do not consistently improve predictive performance for Gemini-3.0-Flash and GPT-5.4 under the conditions examined. This may partly reflect chain-of-thought inference capabilities developed during post-training, potentially reducing the need for prescribed procedures. Accordingly, we use the "No protocol" prompt for subsequent analyses unless otherwise specified. In the Guided mode, model-derived initial estimates are additionally supplied. Because higher sampling temperatures generally increase output variability, we fixed the temperature at zero and assessed reproducibility across three independent runs using identical prompts. As shown in Figs. S4, 5, Gemini-2.5-Flash exhibited greater reproducibility than Gemini-2.5-Pro on both evaluated datasets, although neither model produced fully consistent predictions across all samples, indicating that residual run-to-run variability remains a limitation of LLM-based property prediction even at zero temperature. The experiments presented throughout this study employ different LLMs (Guided or Unguided), reference samples, and repetition protocols. Table 2 summarizes the settings for each experiment, including the specific LLM version, Guided or Unguided mode, prompting protocol, reference samples, and number of repeated runs.

### Assessing potential pretraining leakage with a matched recent cohort

To investigate whether potential memorization of previously reported alloy data contributes to LLM prediction performance, we performed a temporally separated matched-sample analysis using Ti-alloys. A set of 51 recently reported samples was collected from papers published in December 2025 or later. Given their recent publication dates, these samples were considered less likely to have been included in the training data of the evaluated LLMs. Each recently reported sample was paired one-to-one with a previously reported sample from the original dataset based on similarity in alloy composition and processing conditions. The matching procedure used only composition and processing variables, while target property values, paper identities, and sample identifiers were excluded to avoid introducing information leakage. Both sets of matched samples were evaluated using GPT-5.5 under identical prediction conditions, including the same reference/training pool, input representation, prompt configuration, and prediction procedure. Each evaluation was repeated three times using the same 51 matched sample pairs, and MAE was used as the primary performance metric. As shown in Table 3, the UTS MAE was 170.37 ± 6.25 MPa for the previously reported samples and 164.48 ± 11.64 MPa for the recently reported samples, corresponding to a ΔMAE of −5.89 MPa. In contrast, the elongation MAE increased from 3.54 ± 0.20 to 4.88 ± 0.15 percentage points, corresponding to a ΔMAE of +1.34 percentage points. Therefore, previously reported samples achieved lower prediction errors only for elongation but not for UTS, and no consistent performance advantage was observed across the two target properties. These results suggest that even if previously reported alloy data had been encountered by LLMs during pretraining, such prior exposure did not provide a measurable prediction advantage compared with recently reported samples that were less likely to be included in the training data. Under this matched-sample evaluation, we find no evidence that memorization of existing alloy literature substantially improves LLM-based property prediction accuracy.

**Table 3** Matched comparison of GPT-5.5 prediction errors between previously reported and recently reported Ti-alloy samples using full training dataset. MAE values are reported as mean ± standard deviation over three repeated runs using the same 51 matched sample pairs. UTS and elongation errors are expressed in MPa and percentage points, respectively. ΔMAE is calculated as $MAE_{recent}$-$MAE_{original}$; negative values indicate lower prediction errors for recently reported samples, whereas positive values indicate lower prediction errors for previously reported samples.

| Target | Runs | Pairs | Original MAE | Recent MAE | ΔMAE |
|---|---|---|---|---|---|
| UTS (MPa) | 3 | 51 | 170.37 ± 6.25 | 164.48 ± 11.64 | -5.89 |
| El (%) | 3 | 51 | 3.54 ± 0.20 | 4.88 ± 0.15 | +1.34 |

## In-distribution benchmark

To evaluate the predictive capability of LLMs under in-distribution conditions, we first compared GPT-5.4 with several representative baselines, including conventional machine-learning models (XGBoost, LightGBM, Random Forest, CatBoost, and MLP), materials-domain BERT-based approaches (SteelBERT, MatSciBERT, and SciBERT), TabPFN models, and the LLM-Prop framework. For the conventional ML and BERT-based approaches, we report the best-performing model among each model family based on the test performance, while the results of TabPFN-N, TabPFN-T, LLM-Prop, GPT-5.4 and GPT-5.5 are directly reported. TabPFN-N uses the original structured material features as input, whereas TabPFN-T adopts the same textual material descriptions provided to the LLM. As summarized in **错误!不能识别的开关参数。**,

GPT-5.5 achieves the lowest MAE in four out of seven alloy property prediction tasks, including HEA-UTS, Steel-UTS, Ti-El, and Steel-El, demonstrating competitive predictive performance relative to both feature-based machine-learning methods and specialized tabular learning models. This suggests that LLMs can be effective in utilizing heterogeneous material representations, including compositional information and processing descriptions, and capture complex relationships between material characteristics and target properties through their pretrained knowledge and in-context learning.

Table 4 Five-fold random cross-validation (Rand-CV) performance comparison for alloy property prediction tasks using full training dataset. The results are reported as mean absolute error (MAE), represented as mean ± standard deviation (SD) across five random cross-validation folds. For the Trad. and BERT categories, the best-performing model within each model family is reported. Specifically, Trad. represents the best result among XGBoost, LightGBM, Random Forest, CatBoost, and MLP, while BERT represents the best result among SteelBERT, MatSciBERT, and SciBERT with downstream neural networks. TabPFN-N, TabPFN-T, LLM-Prop, GPT-5.4 and GPT-5.5 report their individual model performance. Lower MAE values indicate better prediction accuracy, and the best performance for each task is highlighted in bold.

| Task | Trad. | BERT | TabPFN-N | TabPFN-T | LLM-Prop | GPT-5.4 | GPT-5.5 |
|---|---|---|---|---|---|---|---|
| Al-UTS | 29.50 ± 2.39 | 38.97 ± 2.96 | 27.89 ± 2.37 | 30.13± 2.53 | 42.00 ± 4.21 | 31.15±2.86 | 33.65 ± 1.87 |
| Ti-UTS | 78.51 ± 15.21 | 118.03 ± 15.34 | 71.16 ± 15.70 | 93.74±10.83 | 125.69 ± 11.87 | 74.80±15.83 | 84.14 ± 17.27 |
| HEA-UTS | 113.91±11.47 | 168.03 ± 34.32 | 107.26 ± 8.93 | 134.20±14.12 | 184.89 ± 4.97 | 103.25 ± 7.22 | 121.77 ± 13.57 |
| Steel-UTS | 136.39 ± 9.37 | 150.77 ± 19.96 | 132.00 ± 12.22 | 114.96±11.53 | 153.28 ± 17.69 | 80.40 ± 12.43 | 110.36 ± 10.37 |
| **Ti-El** | **3.89 ± 0.61** | **5.52 ± 0.78** | **3.92 ± 0.44** | **4.44 ± 0.52** | **5.93 ± 0.76** | **3.58 ± 0.37** | **3.83 ± 0.46** |
| **HEA-El** | **8.61 ± 0.59** | **12.52 ± 2.60** | **8.60 ± 0.54** | **9.69 ± 0.74** | **11.60 ± 0.46** | **8.07 ± 1.04** | **8.95 ± 0.79** |
| **Steel-El** | **7.24 ± 0.95** | **6.96 ± 1.54** | **7.17 ± 0.91** | **5.95 ± 0.91** | **7.53 ± 0.79** | **4.91 ± 1.06** | **5.64 ± 1.09** |
| **Matbench Steel-YS** | **91.28 ± 22.19** | **130.91 ± 15.79** | **76.64 ± 10.54** | **75.46 ± 11.85** | **150.87 ± 21.23** | **91.56 ± 8.98** | **93.86 ± 15.17** |

## Matbench comparison

To further evaluate the approach beyond the four alloy datasets examined in this study, we additionally assessed Gemini-3-Flash on two regression tasks from the Matbench benchmark[41] using the unguided mode: the steels and jdft2d datasets. The former consists of 312 samples with compositions and target property yield strength, while the latter consists of 636 samples with crystal structure and target property DFT-calculated exfoliation energy. The results were compared with representative composition-based and structure-based models from the Matbench leaderboard. For steels, as shown in Table 5, GPT-5.4 under the No-protocol setting yielded an MAE of 93.65 ± 13.15, close to that obtained with the full prompt. This small difference again suggests that advanced LLMs appear relatively insensitive to details of prompt-protocols. In addition, Gemini-3-Flash achieved a Mean Absolute Error (MAE) of 98.08 ± 10.72, comparable to GPT-5.4, which reached 95.64 ± 15.72. Notably, despite relying solely on few-shot prompting without task-specific training, it substantially outperformed GPTChem[48](143.00 ± 18.97), which was explicitly fine-tuned on this dataset. GPT-5.5 achieved a Mean Absolute Error (MAE) of 82.83±9.64, which is better than MODNet[46] (87.76 ± 13.66) and comparable to TPOT-Mat[47] (79.95 ± 15.19).

For JDFF2D (Table 6), we utilized Robocrystallographer[60] as in LLM4MatBench[35]—a tool that generates deterministic, human-readable text descriptions of crystal structures from their CIF files—to convert structural data into textual prompts for the LLM. Gemini-3-Flash achieved an MAE of 33.81 ± 10.83, surpassing both CGCNN[65] (49.24 ± 12.95) and coNGN[66] (36.17 ± 12.97), and closely approaching the performance of MODNet[67] (33.19 ± 8.21). GPT-5.4 and GPT-5.5 also showed competitive performance, with an MAE of 36.21 ± 9.73 and 36.29±11.43. These results are consistent with the conclusion of Ref.[34] with LLM-Prop, that when crystal structures are represented through physically meaningful textual descriptions, LLMs can perform accurate materials property regression without explicit graph-based architectures or task-specific model training. In addition, Gemini-3-Flash outperforms specialized GNN models such as coNGN, CGCNN on jdft2d, thus LLM-based text inference appears not merely a convenient alternative representation strategy, but could serve as a genuinely competitive predictive paradigm for materials science.

**Table 5** Performance of Gemini-3-Flash Preview, GPT-5.4, GPT-5.5 and representative Matbench leaderboard models on the steels task (312 samples) using full training dataset. Gemini-3-Flash Preview and GPT-5.4 were evaluated using the full prompt. MODNet, TPOT-Mat and GPTChem results are taken from the Matbench leaderboard. An additional No-protocol ablation was conducted for GPT-5.4, giving an MAE of 93.65 ± 13.15, comparable to its full prompt MAE of 95.64 ± 15.72.

| Model | MAE | RMSE | MAPE | Max Error |
|---|---|---|---|---|
| **Gemini-3-Flash Preview (Full prompt)** | **98.08± 10.72** | **153.27±30.69** | **0.07±0.01** | **678.56±281.63** |
| **GPT-5.4 (No protocol)** | **93.65 ± 13.15** | **139.01± 26.19** | **0.07±0.01** | **598.66±171.32** |
| **GPT-5.4 (Full prompt)** | **95.64± 15.72** | **143.96± 34.25** | **6.78±0.72** | **625.04±282.50** |
| **GPT-5.5** | **82.83±9.64** | **122.53±18.38** | **0.06±0.01** | **449.41±96.23** |
| **MODNet[67]** | **87.76± 13.66** | **144.77± 41.87** | **0.06±0.01** | **722.36±309.64** |
| **TPOT-Mat[47]** | **79.95± 15.19** | **115.37± 25.11** | **0.06±0.01** | **407.62 ± 95.24** |
| **GPTChem[48]** | **143.00±18.97** | **218.03± 31.92** | **0.10±0.01** | **906.16±328.70** |

**Table 6** Performance of Gemini-3-Flash Preview, GPT-5.4, GPT-5.5 and representative Matbench leaderboard models on the JDFT2D task (636 samples) using full training dataset. The matbench_jdft2d dataset uses crystal structures of two-dimensional materials as inputs, with the target property being the DFT-calculated exfoliation energy, which evaluates the predicted ease of exfoliating layered materials.

| Model | MAE | RMSE | MAPE | Maximum Error |
|---|---|---|---|---|
| **Gemini-3-Flash Preview (Full prompt)** | **33.81±10.83** | **95.97±49.01** | **5.47 ± 11.37** | **703.73±516.80** |
| **GPT-5.4 (Full prompt)** | **36.21 ± 9.73** | **102.37±44.52** | **5.53±11.43** | **727.51± 446.96** |
| **GPT-5.5** | **36.29±11.43** | **101.52±47.33** | **5.57±11.49** | **697.02±442.37** |
| **CGCNN[65]** | **49.24±12.95** | **112.77±53.91** | **7.18 ± 14.86** | **709.24± 556.11** |
| **MODNet[67]** | **33.19 ± 8.21** | **96.73 ± 45.13** | **4.02 ± 8.18** | **721.82± 518.81** |
| coNGN[66] | 36.17±12.97 | 95.48 ± 53.22 | 4.70 ± 9.58 | 660.12± 528.84 |

## Similarity-based retrieval and reference ratio

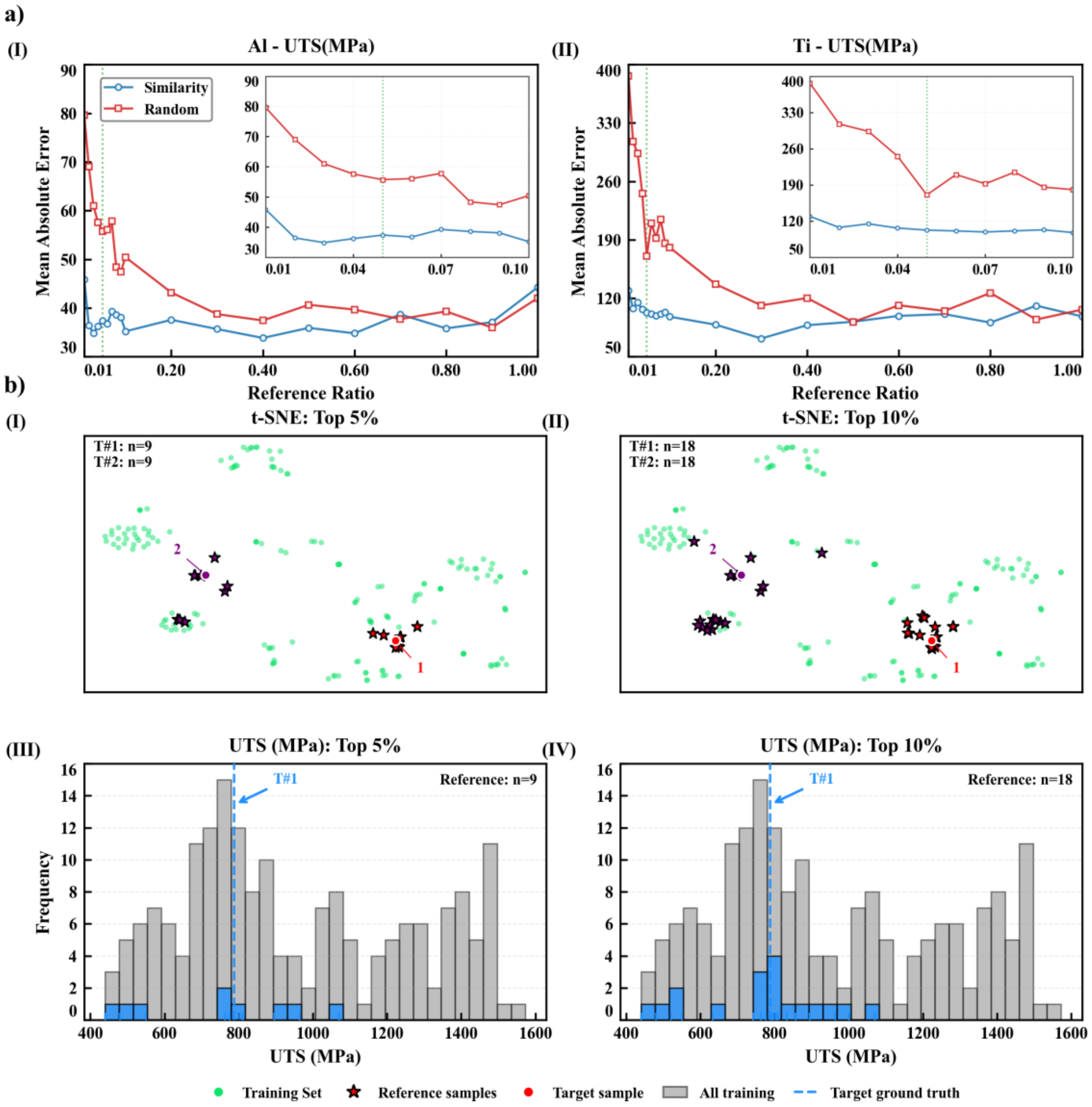

**Figure 2** Effect of reference-sample selection and reference-set size on UTS prediction. a) Prediction performance for Al and Ti alloy UTS using semantic similarity-based retrieval and random sampling. Panels I, II show MAE for Al and Ti alloys, respectively. Insets show the low ratio range [0.01-0.1]. The green line marks 5% reference ratio. Similarity-based retrieval near-saturated performance with small fraction of high-relevance reference samples, b) Panels I, II show t-SNE projections of retrieved Top 5% and Top 10% reference samples for Ti target samples. Panels III and IV show UTS distributions of reference samples compared with full training set. Green denotes training samples, colored stars target samples and their retrieved references in the embedding space, gray denote full training-set distribution, and blue denote the retrieved reference samples. The blue dashed line is the ground-truth UTS of target sample T#1.

To examine whether LLM-based property prediction requires a large number of contextual examples, we investigated the effect of reference-sample selection and reference-set size on UTS prediction for Al and Ti alloys. The prompt was fixed to the complete full prompt, including establishing baseline, plausibility assessment and interpolative correction, so that the effect of data selection could be isolated from prompt design. Two strategies were compared: semantic similarity-based retrieval using the all-MiniLM-L6-v2 embedding model, and random sampling from the training set. This comparison evaluates whether LLM prediction is mainly governed by the quantity of contextual samples or by the quality and relevance of the retrieved references.

As shown in Figure 2a, semantic similarity-based retrieval exhibits a clear quality-dominant advantage. By selecting the Top-n% most relevant samples according to cosine similarity in the learned embedding space, the method rapidly reaches high prediction accuracy using only a small fraction of the training data. At a reference ratio of approximately 5%, similarity-based retrieval already achieves near-saturated performance, with $R^2$>0.8 for Ti alloys and MAE below 40 MPa for Al alloys. Notably, this 5% high-quality reference set yields performance comparable to, and in some cases better than, using the entire training set. This indicates that a small number of semantically and physically relevant samples can effectively replace the brute-force use of all available data. In contrast, random sampling is much less efficient. Increasing the number of randomly selected samples leads to slower and less stable improvement, because many added samples are weakly related to the target alloy system or processing route. Even when the full training set is used, random sampling does not consistently outperform similarity-based retrieval with only 5% of the data, revealing an approximately 20-fold efficiency gap.

The origin of this advantage is illustrated in Figure 2b and Figs. S9-11. The t-SNE projections show that the all-MiniLM-L6-v2 embeddings organize semantically related alloys into local neighborhoods. The Top 5% and Top 10% retrieved samples cluster around their corresponding target samples rather than dispersing randomly, indicating that the embedding distance captures latent similarities in composition and processing. This locality provides the LLM with a compact and target-relevant evidence base for comparison and interpolation. The retrieved samples also show stronger alignment in property space. As shown by the UTS histograms in Figure 2b (III, IV) and Figs. S9-11, the full training set spans a broad and multimodal property distribution, whereas the retrieved references occupy a narrower range closer to the target property. For target sample T#1, the retrieved UTS distribution is closely aligned with the ground-truth value. Thus, similarity-based retrieval acts as a physics-aware filter that removes irrelevant statistical noise while retaining samples with analogous mechanical behavior.

For the Al alloy dataset, when 20% of the training samples (132 references) are used as shown in Figure 2a, similarity-based selection achieves a high average cosine similarity of 0.970, with the lowest similarity among selected references still reaching 0.959. However, across 1,000 randomly sampled reference sets of the same size, 99.98% contain at least one reference whose similarity exceeds the average similarity of the similarity-selected set. This indicates that the Al dataset contains highly dense local neighborhoods, allowing random subsets to frequently recover informative references without explicit similarity-based retrieval. This observation is consistent with the intrinsic characteristics of the dataset: the Al dataset consists exclusively of 7xxx-series Al alloys with relatively narrow composition and processing distributions, and 83.8% of test samples have at least one training neighbor with cosine similarity ≥ 0.99.

Nevertheless, data redundancy alone cannot fully account for the observed LLM performance. As shown in Table 10, even when k-NN is optimized across different reference sizes, representation spaces, and aggregation strategies, GPT-5.5 maintains comparable or superior performance in most tasks. Specifically, using the same reference samples as the optimized k-NN baseline, GPT-5.5 achieves lower MAE in 7 out of 8 tasks. These results suggest that the convergence between random and similarity-based selection in the Al dataset is primarily driven by high data redundancy, whereas the remaining performance advantage of LLMs likely arises from their ability to integrate information across multiple reference examples rather than relying solely on direct nearest-neighbor matching. Furthermore, the results in Figs. S12-16, obtained using a fixed random seed, show that general-purpose LLMs can outperform domain-specific BERT-based models through few-shot prompting, particularly in predicting alloy elongation. These results underscore the potential of pretrained LLMs to achieve accurate property predictions from composition and processing using a limited set of reference examples, without domain-specific fine-tuning or task-specific downstream network design.

**Guided vs Unguided**

To assess the effectiveness of the "Guided mode" on the reference samples, we compared the Unguided mode across alloy prediction tasks and reference ratios, using a fixed random seed to split the data (Figs. S17-25). The Guided mode reduced MAE in most tasks, with particularly pronounced benefits at low reference ratios (0–10%), where an additional baseline estimate from the best-performing prior model for the corresponding task was supplied as a numerical anchor (Figs. S6–8). These benefits diminished as more reference samples were included, suggesting that prior estimates complement limited local evidence. In contrast, the Steel dataset showed no consistent improvement, potentially owing to substantial variability in processing conditions and input descriptions that were not derived from standardized structured data. For such inputs, LLMs may draw on semantic associations amongst reference samples to identify processing–property relationships that are not fully captured by explicit features. If the prior model inadequately captures these relationships, biases in its predictions may limit the benefits of guidance. Detailed experimental results and analyses are provided in Figs. S17, 18.

### Influence of representation space on LLM and k-NN prediction performance

To investigate the influence of reference representation space on prediction performance, we systematically compared four material representations for the Ti-UTS prediction task (Fig. S26 and Table 7), including the original standardized numerical features, text embeddings derived from LLM input representations, and two-dimensional projections obtained using principal component analysis (PCA-2D) and uniform manifold approximation and projection (UMAP-2D). Under a fixed reference budget of 5% of the training set, the corresponding k-NN baseline achieved MAEs of 100.2, 125.8, 123.1, and 173.9 MPa, respectively. Among these representations, numerical features provided the most effective neighborhood structure for k-NN prediction, whereas UMAP-2D resulted in the largest error, indicating that the choice of representation strongly influences similarity-based reference retrieval.

In contrast, GPT-5.5 exhibited a different representation preference, achieving the lowest MAE with PCA-2D (77.5 MPa), followed by text embeddings (78.3 MPa), numerical features (80.6 MPa), and UMAP-2D (100.5 MPa). This difference suggests that the optimal representation for LLM-based prediction does not necessarily coincide with that for conventional k-NN regression, likely because LLMs can exploit higher-order relationships contained in the retrieved reference samples beyond direct feature-space similarity. Despite the variation among representations, GPT-5.5 consistently outperformed the corresponding k-NN baselines using identical reference samples, achieving relative MAE reductions ranging from 19.6% to 42.2%. These results demonstrate that while representation selection affects both retrieval quality and prediction accuracy, the performance advantage of LLM-based prediction over k-NN is robust across different material representations rather than being limited to a specific feature space.

**Table 7** Comparison of LLM and k-NN prediction performance across reference representations for the Ti-UTS task with the reference ratio fixed at 5% of the training set. Reference samples were retrieved using standardized numerical features, text embeddings, PCA-2D, or UMAP-2D, with the reference ratio fixed at 5% of the training set. Within each representation, the LLMs and the corresponding k-NN baseline used identical reference samples. MAE is reported in MPa, with lower values indicating better performance. Relative MAE reduction is calculated as $(MAE_{k\text{-}NN}-MAE_{LLM})/MAE_{k\text{-}NN}\times 100\%$ . GPT-5.5 MAEs are averaged over three repeated runs

| Representation | k-NN | GPT-5.5 | Reduction (%) | SD over repeats (%) | 95 % interval, run01 | 95 % interval, run02 | 95 % interval, run03 |
|---|---|---|---|---|---|---|---|
| **Numeric z (cosine)** | **100.2** | **80.6** | **19.6%** | **2.8%** | **2.8%** | **[9.3%, 30.6%]** | **[9.9%, 32.1%]** |
| **Text embedding** | **125.8** | **78.3** | **37.8%** | **3.0%** | **3.0%** | **[27.4%, 46.4%]** | **[31.4%, 49.6%]** |
| **PCA-2D** | **123.1** | **77.5** | **37.0%** | **1.4%** | **1.4%** | **[25.9%, 44.1%]** | **[28.5%, 46.2%]** |
| **UMAP-2D** | **173.9** | **100.5** | **42.2%** | **0.3%** | **0.3%** | **[33.3%, 49.8%]** | **[33.2%, 50.3%]** |

## LLM versus k-NN at a fixed reference budget

Since both LLM-based few-shot prediction and k-NN regression retrieve similar samples from the training set, we further examined whether the predictive gains of LLMs could be attributed to similarity-based retrieval alone. For a controlled comparison, both approaches were evaluated at a fixed reference budget of 5% of the training set. For the k-NN baseline, we systematically assessed four material representations, namely numeric, text, PCA-2D, and UMAP-2D. Each representation was evaluated using both uniform averaging and similarity-weighted averaging. The best-performing combination of representation and aggregation strategy was designated as the optimized k-NN at 5% reference budget baseline and compared with Gemini-2.5-Pro, Gemini-2.5-Flash, and GPT-5.5. (Table 8).

To evaluate whether LLMs can provide predictive advantages beyond conventional similarity-based retrieval, GPT-5.5 was first compared with optimized k-NN baselines using a fixed 5% reference budget. As shown in Table 8, GPT-5.5 consistently outperformed the corresponding k-NN models across all eight alloy property prediction tasks, achieving MAE reductions of 23.78–34.35%. The largest improvements were observed for Ti-El (34.35%), Al-UTS (32.12%), Steel-UTS (31.85%), and HEA-UTS (31.57%). These results demonstrate that, even when provided with only a small subset of reference samples, GPT-5.5 can effectively extract predictive information beyond direct similarity matching, indicating that LLM-based prediction benefits from integrating retrieved examples with broader contextual material knowledge.

To further examine whether the observed improvement originates from reference sample selection or from the intrinsic predictive capability of the LLM, a stricter matched-reference comparison was performed. The k-NN configuration was further optimized by searching over reference size, material representation, and aggregation strategy, and GPT-5.5 was evaluated using exactly the same reference samples selected for the optimized k-NN model. As summarized in Table 9, GPT-5.5 achieved lower MAE than the optimized k-NN baseline in seven out of eight tasks. Significant improvements were obtained for Steel-UTS (29.56%), HEA-UTS (26.32%), HEA-El (22.43%), and Steel-El (21.76%). Only Al-UTS exhibited slightly higher error than k-NN, with a 2.94% increase in MAE. These results indicate that the advantage of LLM-based prediction is not solely derived from retrieving similar samples, but also from its ability to interpret and combine reference information with learned material representations, enabling more effective prediction with limited reference data.

**Table 8** Performance comparison between LLM-based prediction and k-NN baselines under a fixed 5% reference budget. MAE reduction is calculated relative to the corresponding k-NN baseline: $(MAE_{k\text{-}NN}\text{-}MAE_{LLM})/\mathrm{MA}E_{k\text{-}NN}\times 100\%$

| Task | k at 5 % | Optimized k-NN at 5 %: MAE (representation, weighting) | Gemini-2.5-Pro MAE | Gemini-2.5-Flash MAE | GPT-5.5 MAE | Reduction, Pro | Reduction, Flash | Reduction, GPT-5.5 |
|---|---|---|---|---|---|---|---|---|
| Al-UTS | 31 | 74.51 (numeric, weighted) | 62.83 | 63.08 | 50.58 | 15.68 | 15.33 | 32.12 |
| HEA-El | 19 | 18.90 (text, weighted) | 13.01 | 12.24 | 13.62 | 31.18 | 35.25 | 27.94 |
| HEA-UTS | 19 | 252.42 (numeric, weighted) | 183.97 | 175.04 | 172.73 | 27.12 | 30.66 | 31.57 |
| Steel-El | 27 | 14.20 (numeric, weighted) | 9.33 | 10.39 | 9.87 | 34.28 | 26.79 | 30.49 |
| Steel-UTS | 27 | 249.92 (numeric, weighted) | 163.96 | 172.01 | 170.32 | 34.40 | 31.17 | 31.85 |
| Ti-El | 9 | 7.81 (numeric, weighted) | 6.71 | 6.98 | 5.13 | 14.09 | 10.64 | 34.35 |
| Ti-UTS | 9 | 150.82 (numeric, weighted) | 126.99 | 153.44 | 114.96 | 15.80 | −1.74 | 23.78 |
| Matbench Steel-YS | 12 | 102.96 (numeric, weighted) | not available | not available | 78.35 | not available | not available | 23.90 |

**Table 9** Comparison of GPT-5.5 prediction performance with optimized k-NN baselines under matched reference conditions. The table compares GPT-5.5 with the best-performing k-NN baseline obtained using the same reference samples. For each task, the k-NN baseline is optimized over different representations, numbers of nearest neighbors (k), and weighting strategies, with the selected configuration reported together with its MAE. GPT-5.5 receives the identical reference set as the corresponding k-NN baseline, and the reported MAE reduction is calculated as $(\mathrm{MAE}_{\text{k-NN}}\text{-}\mathrm{MAE}_{\text{GPT-5.5}})/\mathrm{MAE}_{\text{k-NN}}\times 100\%$. GPT-5.5 results are averaged over three independent runs with identical inputs; SD across repeats and 95% confidence intervals

for each run are provided to quantify stochastic variability.

| Task | Optimized k-NN: MAE (representation, k, weighting) | GPT-5.5 MAE (matched references) | Reduction vs k-NN (%) | SD over repeats (%) | 95 % interval, run01 | 95 % interval, run02 | 95 % interval, run03 |
|---|---|---|---|---|---|---|---|
| Al-UTS | 38.26 (text, 2, weighted) | 39.38 | -2.94 | 0.45% | [-5.92, 8.54] | [-5.46, 8.20] | [-6.11, 8.18] |
| HEA-El | 10.81 (numeric, 2, weighted) | 8.38 | 22.43 | 1.31% | [-3.96, -1.08] | [-3.68, -0.94] | [-4.08, -0.96] |
| HEA-UTS | 146.43 (numeric, 3, uniform) | 107.88 | 26.32 | 0.49% | [-61.23, -14.53] | [-62.23, -15.44] | [-63.63, -14.11] |
| Steel-El | 7.21 (text, 2, weighted) | 5.64 | 21.76 | 2.38% | [-2.53, -0.52] | [-2.81, -0.66] | [-2.55, -0.39] |
| Steel-UTS | 128.51 (text, 2, weighted) | 90.52 | 29.56 | 5.05% | [-64.06, -23.14] | [-60.18, -14.40] | [-57.49, -11.71] |
| Ti-El | 4.32 (numeric, 3, uniform) | 3.80 | 12.01 | 4.45% | [-1.60, 0.11] | [-1.49, 0.52] | [-1.26, 0.46] |
| Ti-UTS | 85.71 (numeric, 1, weighted) | 78.66 | 8.22 | 2.74% | [-14.66, 7.06] | [-18.73, 0.86] | [-17.98, 6.50] |
| Matbench Steel-YS | 82.54 (numeric, 3, weighted) | 77.46 | 6.16 | 2.07% | [-24.28, 9.84] | [-22.20, 10.35] | [-17.86, 10.66] |

## Optimizing the reference-set size

To investigate whether a compact set of relevant reference samples could replace the full training set as LLM context and reduce the computational cost of subsequent OOD analyses, we systematically evaluated the effect of reference sample size on LLM prediction performance. Each dataset was first randomly partitioned using a fixed random seed of 42. We then optimized the reference retrieval strategy by evaluating k-NN performance across four material representations and two aggregation strategies with varying neighborhood sizes. The optimal representation and corresponding neighborhood size ($k_{\text{k-NN-opt}}$) identified from this analysis were fixed for each task. Subsequently, reference samples for LLM prediction were selected according to similarity ranking within the optimized representation, and the number of retrieved reference samples was varied from 1 to $k_{\text{k-NN-opt}}$ + 5 to evaluate the dependence of LLM performance on reference set size.

As shown in Figure 3 and Fig. S27, LLM prediction performance generally stabilized using only a small number of similarity-selected reference samples, indicating that full training-set context is not necessary for accurate prediction. Across seven of eight tasks, the LLM achieved comparable or lower MAE than the best-performing k-NN baseline when evaluated around the corresponding k-NN-optimal reference size. In several tasks, only a few highly relevant reference samples were sufficient to achieve prediction errors close to those obtained using the full reference set, while substantially reducing the amount of contextual information provided to the LLM. Although some tasks, such as HEA-UTS and HEA-YS, continued to benefit from additional reference samples, the overall results demonstrate that the k-NN-optimized retrieval strategy provides an effective approach for identifying compact reference sets. These findings suggest that subsequent OOD analyses can avoid using the complete training set as LLM context and instead adopt optimized low-budget reference sets to reduce input length and inference cost while preserving the major prediction trends.

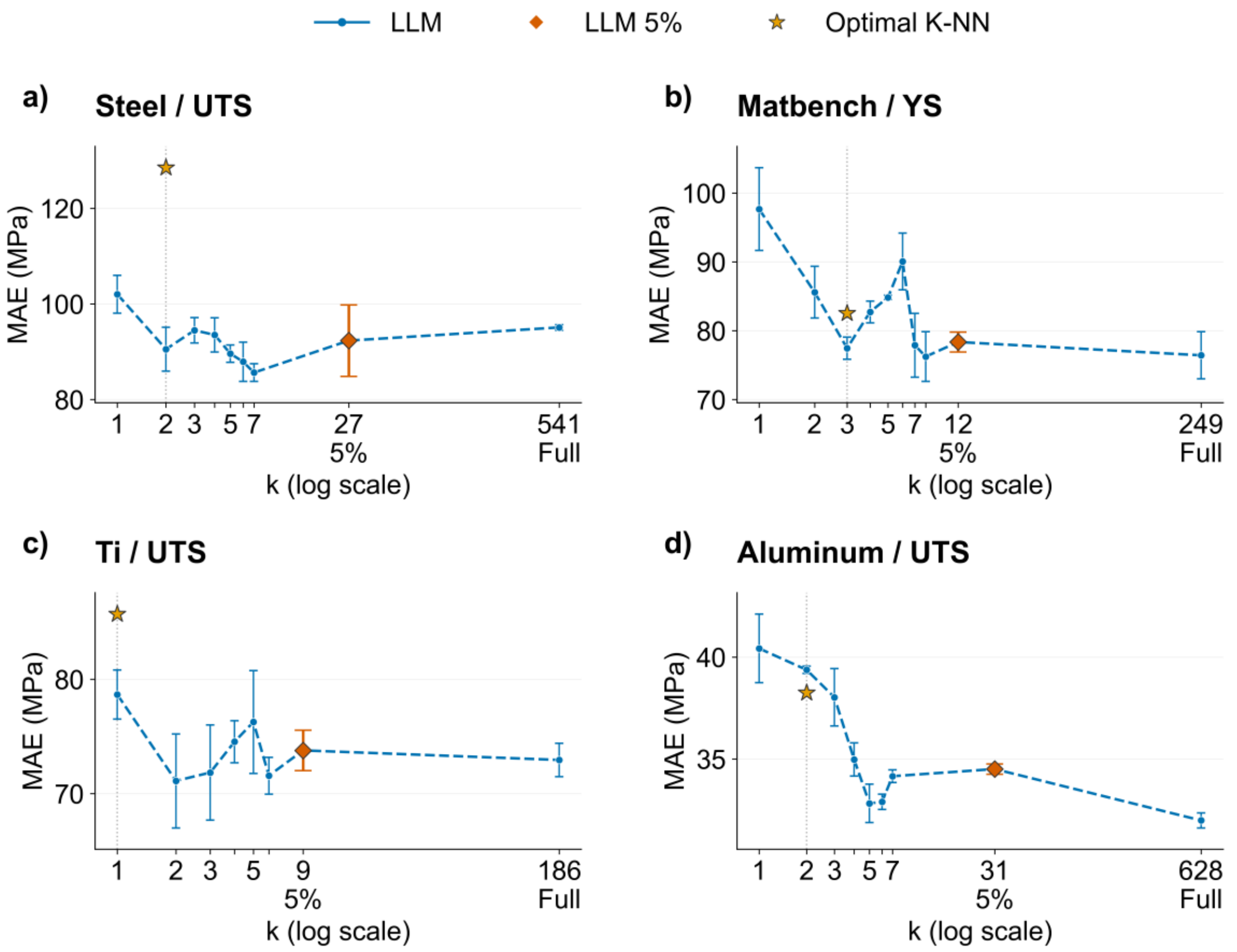

**Figure 3** Effect of reference sample count on LLM prediction performance and comparison with the best-performing k-NN baseline. Panels a–d show the results for Steel-UTS, Matbench-YS, Ti-UTS, and Al-UTS, respectively. Blue circles and connecting lines indicate LLM predictions at different reference sample counts $k$, with $k$ varied from 1 to the corresponding k-NN-optimal $k + 5$; orange diamonds mark the results at a 5% reference budget, and yellow stars indicate the best-performing k-NN baseline. Gray vertical dotted lines mark the corresponding k-NN-optimal $k$. The horizontal axis uses a logarithmic scale, and MAE is reported in MPa. Full denotes the use of all training samples as references. Evaluations were repeated three times for Steel, HEA, Matbench, Ti and Al. The percentages below each panel indicate the relative change in MAE at the 5% reference budget compared with the full-reference setting. Results for all eight tasks are provided in Fig. S27.

**OOD data and Wasserstein distance**

Materials property prediction in real-world alloy discovery is intrinsically an out-of-distribution (OOD) problem because the compositions, processing histories, and high-value property regions of interest often lie outside the dense regions covered by available experimental data. We benchmarked all model families under one Random-CV baseline and six OOD protocols using the unguided mode: extrapolation, leave-one-cluster-out (LOCO)[51], SparseXsingle(SX-sgl), SparseYsingle(SY-sgl), SparseXcluster(SX-cls), and SparseYcluster(SY-cls)[52] The distributions for the titanium alloy UTS datasets and their different target sets are shown in Figure 4. We evaluated the influence of the number of clusters on LOCO CV performance by varying the number of clusters (k) from 2 to 10. As shown in Fig. S35, the Pearson R values remain relatively stable across different k values for different benchmark datasets and ML methods, indicating that the conclusions are not sensitive to the choice of cluster number. We selected k = 5 as a balanced setting that provides sufficient separation of the feature space while maintaining a reasonable number of samples within each cluster for OOD evaluation. Additionally, Figs. S36-42 provide visualizations for the aluminum alloy, high-entropy alloy, steels and Matbench steel yield strength datasets. In Random-CV, held-out samples are distributed across the same dense regions as the training data, so the evaluation mainly measures interpolation. In contrast, extrapolation isolates high-property tails, LOCO removes entire feature-space clusters, SparseX selects low-density regions in the alloy representation space, and SparseY selects rare regions of the property distribution. The sample size of each target set are as shown in Table 1.

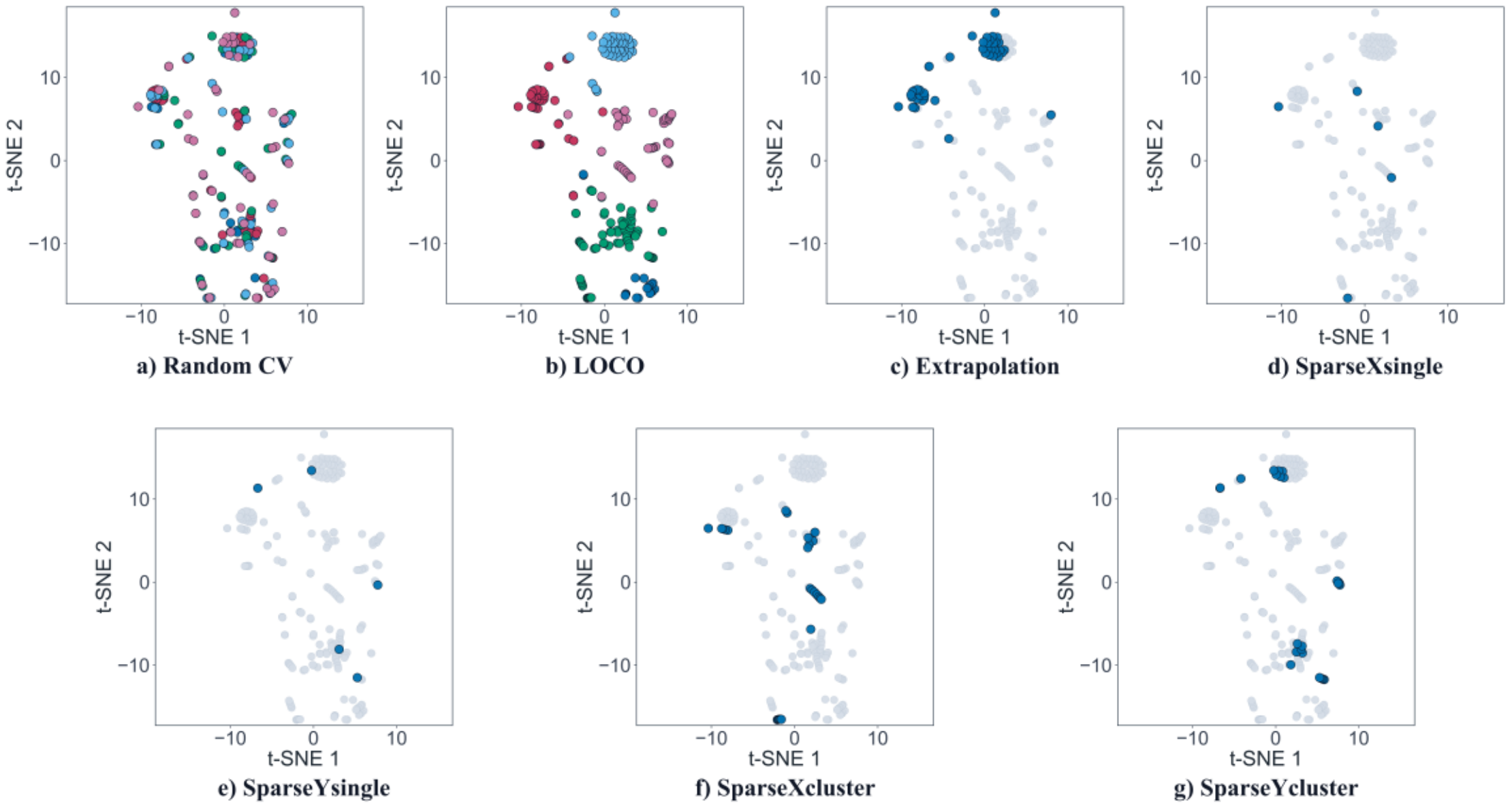

**Figure 4** t-SNE plots for standard cross-validation (CV) test set and six out-of-distribution (OOD) test sets generated using different target selection strategies for Ti alloy ultimate tensile strength (UTS) dataset. Each circle represents one of the 233 samples. (a) Five-fold CV with random splitting, where samples are assigned to 5 color-coded test subsets. (b) Leave-one-cluster-out (LOCO) splitting based on clustered sample groups. (c) Extrapolation-based test set distribution. (d) SparseXsingle, where 5 individual test samples are highlighted in distinct colors and the remaining samples are shown in gray. (e) SparseYsingle, where 5 individual test samples are highlighted in distinct colors and the remaining samples are shown in gray. (f) SparseXcluster, where 29 test clusters are highlighted in distinct colors and the remaining samples are shown in gray. (g) SparseYcluster, where 30 test clusters are highlighted in distinct colors and the remaining samples are shown in gray. The figure illustrates how different target-generation methods vary in their coverage of the feature space and in their degree of distributional shift from the full dataset.

Conventional random cross-validation primarily evaluates interpolation within the observed data manifold and may therefore overestimate model reliability in discovery-oriented screening scenarios. We therefore constructed OOD test sets with deliberate shifts in both target-property and alloy-feature distributions, and used the Wasserstein distance to quantify the severity of each shift relative to the corresponding training distribution. To frame the discussion of OOD generalization, we utilize the mathematical framework of Optimal Transport (OT) theory. At its core, OT seeks the most efficient "transport plan" to transform one probability distribution into another. The resulting metric, the Wasserstein distance, represents the minimum total cost required to move mass (representing data density) from a source distribution $\mu$ (our training alloys) to a target distribution $\nu$ (the novel test systems). In the context of materials informatics, this cost is defined by a ground metric—typically the Euclidean distance within a high-dimensional feature space of chemical compositions and processing parameters—which reflects the physical and chemical dissimilarity between materials. That is,

$$W_p(\mu,\nu)=\left(\inf_{\gamma\in\Gamma(\mu,\nu)}\int_{M\times M} d(x,y)^p \; d\gamma(x,y)\right)^{1/p} \quad (1)$$

Here we focus on the Wasserstein-1 distance ($W_1$), where the ground metric (*d(x,y)*) represents the "cost" of moving a unit of mass from $x$ in the source distribution $\mu$ (the training alloys) to $y$ in the target distribution $\nu$ and the transport plan ($\gamma$) represents the joint distribution (or "coupling") that describes how mass is reassigned[54–57,68]. It must satisfy the marginal constraints:

$$\int_M \gamma(x,y)dy=\mu(x)\text{and}\int_M \gamma(x,y)dx=\nu(y) \quad (2)$$

The Infimum (*inf*) is the distance defined by the *optimal* coupling—the specific plan $\gamma$ that minimizes the total integrated cost of transformation. By quantifying this distance, we can rigorously define the degree of extrapolation required and evaluate how effectively we maintain predictive accuracy moving to increasingly "distant" chemical spaces. To make Wasserstein distances comparable across test splits with different sample sizes and different physical scales, we computed a standardized Wasserstein distance, denoted as $W_1^*$. The details of the calculation are shown in Methods 4.5.

Unlike traditional statistical distances, such as the Mahalanobis or average neighbor distance, the Wasserstein distance does not rely on rigid parametric shapes (like Gaussians or ellipsoids) or just local point-densities, making it exceptionally sensitive to global shifts in data due to its sensitivity to the feature space geometry. Also, the energy distance behaves as a "weak" topology metric, and although similar to a relaxed Wasserstein-1 distance, it lacks strict preservation of horizontal transport geometry. Similarly, the Maximum Mean Discrepancy (MMD) relies entirely on the choice of the kernel (e.g., RBF/Gaussian kernel), and it does not measure distance along a continuous path. Instead, it measures how much the distributions overlap locally based on the kernel's bandwidth, in contrast to optimal transport where a metric is defined. With a standard RBF kernel for MMD, if a test sample is far outside the training distribution, the kernel drops to zero and MMD becomes saturated and loses its ability to recognize how far the OOD the sample is, unlike $W_1$ that scales linearly.

**Benchmarking model robustness under OOD shifts**

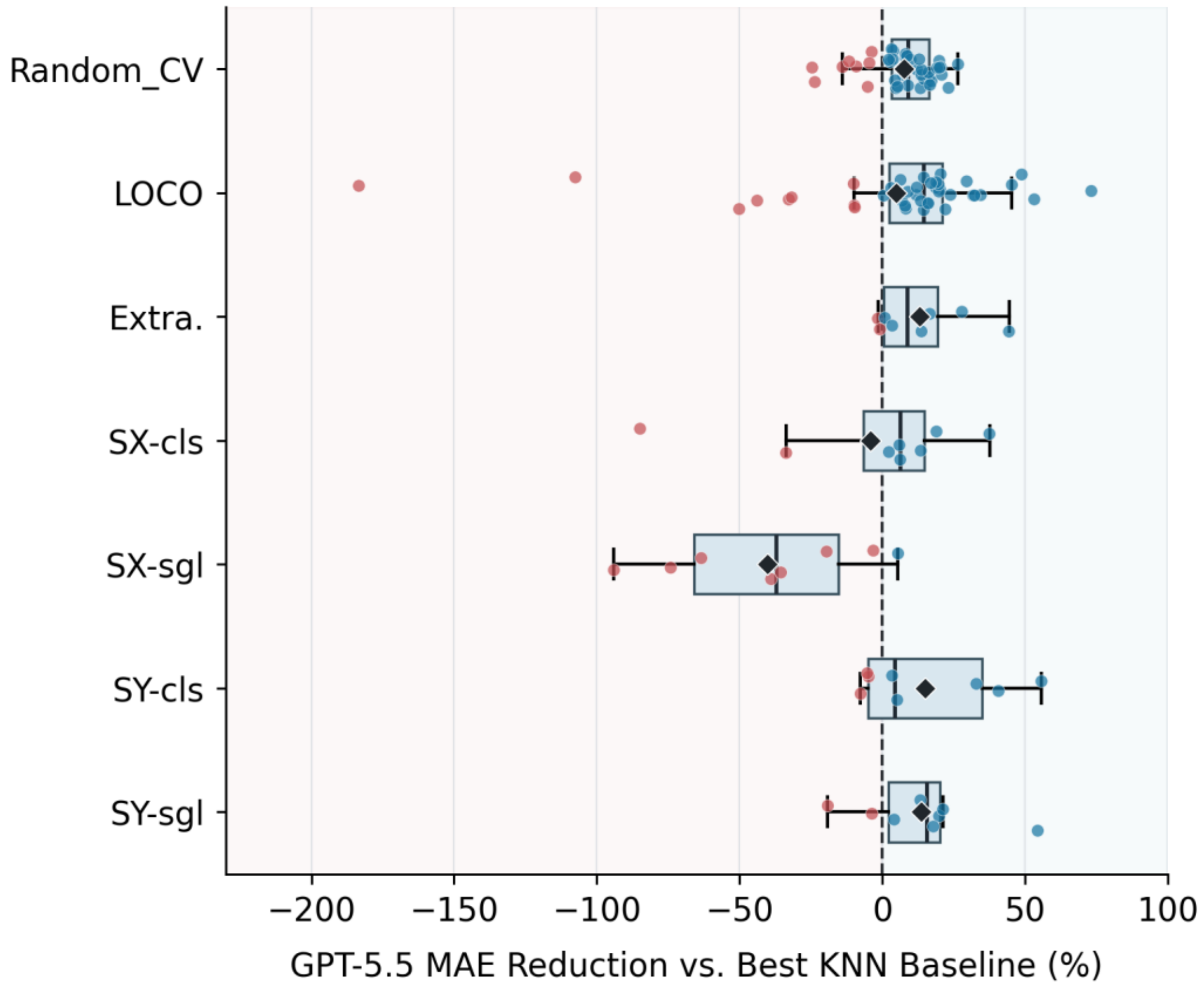

**Figure 5** Comparison of the optimized k-NN predictor and GPT-5.5 with matched k-NN references under OOD conditions. For each task–fold configuration, the KNN model was optimized over four representation spaces, neighborhood sizes *k*, and weighting strategies; the configuration with the lowest test-set MAE was selected. Its exact neighbor set was then provided to GPT-5.5 as few-shot references for the same test samples. Each point represents one configuration; boxes and diamonds show the interquartile range and mean, respectively. The Table 10 shows Relative performance of GPT-5.5 and the selected k-NN predictor across OOD protocols using matched KNN references. GPT-5.5 outperformed k-NN in 87 of 120 configurations, while KNN remained superior in 33.

To assess whether an LLM can make more effective use of retrieved references than direct numerical neighbor regression, we first identified the optimized k-NN configuration for each OOD task–fold as shown in the Figure 5. We searched across text, numerical, PCA-2D, and UMAP-2D representations, as well as different neighborhood sizes k and weighting schemes, and selected the configuration with the lowest test-set MAE. GPT-5.5 was then provided with exactly the same neighbor set and representation used by the selected k-NN model and evaluated on the same test samples.

As shown in Table 7, GPT-5.5 with the matched k-NN references achieved lower MAE than the selected k-NN predictor in 87 of 120 configurations (72.5%), whereas k-NN performed better in the remaining 46 configurations (27.5%). Performance varied substantially across OOD protocols: LOCO produced the highest proportion of GPT-5.5 improvements, while sparse-X single was the least favorable setting, with k-NN outperforming GPT-5.5 in 87.5% of configurations. These results show that retrieved neighbors can provide useful information to the LLM, but GPT-5.5 does not consistently convert that information into more accurate numerical predictions than direct k-NN regression.

**Table 10** Relative MAE performance of GPT-5.5 and the optimized k-NN predictor across OOD protocols using matched KNN references. For each task–fold configuration, a configuration is counted as favorable to GPT-5.5 when it achieves a lower test-set MAE than k-NN; otherwise, it is counted as favorable to k-NN. The final column reports the proportion of configurations in which GPT-5.5 outperforms k-NN within each OOD protocol. Overall, GPT-5.5 achieved lower MAE in 87 of 120 configurations (72.5%), while k-NN performed better in the remaining 33 configurations (27.5%).

| OOD protocol | GPT-5.5 better | k-NN better | GPT-5.5 win rate (%) |
|---|---|---|---|
| Random CV | 32 | 8 | 80.0% |
| LOCO | 31 | 9 | 77.5% |
| Extrapolation | 6 | 2 | 75.0% |
| Sparse-X cluster | 6 | 2 | 75.0% |
| Sparse-X single | 1 | 7 | 12.5% |
| Sparse-Y cluster | 5 | 3 | 62.5% |
| Sparse-Y single | 6 | 2 | 75.0% |
| Overall | 87 | 33 | 72.5% |

In this framework, a small Wasserstein distance indicates that the test samples remain close to the training manifold, whereas a large Wasserstein distance indicates a stronger distributional shift and therefore a more severe extrapolation challenge. This allows the OOD degree of different splitting strategies, alloy systems, and target properties to be compared on a common quantitative basis rather than judged only from visual inspection of two-dimensional projections. The standardized Wasserstein distances are summarized in Table S2, which quantifies the magnitude of distribution shifts in both target space, $W_1^*(y)$, and representation space $W_1^*(x)$. Some target sets are severe primarily in $W_1^*(y)$, whereas others are severe primarily in $W_1^*(x)$, confirming that OOD evaluation does not represent a single difficulty level but rather a family of distinct extrapolation challenges.

The performance comparison for Ti-alloy UTS in Figure 6, together with the results for other datasets in Figs. S43–S49 and the average rankings in Table 11, demonstrate that OOD robustness is highly dependent on both the prediction task and the splitting protocol. No single model family consistently dominates across all alloy systems or target properties. For each dataset split, models were ranked according to their MAE, with rank 1 assigned to the model with the lowest MAE and rank 5 assigned to the model with the highest MAE. The average ranking of model (m) across (N) splits was then calculated as $\overline{R_m} = \frac{1}{N}\sum_{i=1}^{N} R_{m,i}$, where ($R_{m,i}$) denotes the MAE-based rank of model (m) in split (i). Based on this within-split ranking strategy, there is no universal performer. GPT-5.5 achieves the best average rank for HEA-UTS, Steel-UTS, Ti-El and Steel-El, whereas BERT leads on Matbench Steel-YS. Traditional machine-learning models also remain highly competitive for HEA-El.

The task-wise summaries in Figs. S43-49 further corroborate this observation, demonstrating that models with strong performance under random splits are not necessarily optimal under OOD evaluation. This indicates that model ranking is governed not only by the alloy system and target property, but also by the nature of the distribution shift used to define the test set. Random splitting primarily probes interpolation within the observed data distribution. In contrast, extrapolation evaluates global property-space generalization by withholding an extreme region of the target distribution, whereas LOCO assesses cluster-level transfer by excluding an entire region of the composition and processing feature space from training. The SparseX and SparseY protocols provide complementary density based OOD settings. SparseX selects test samples from low-density regions of the input-feature space, representing rare composition and processing conditions, while SparseY selects samples from low-density regions of the target-property space, representing rare or extreme property values. Their cluster variants further evaluate local sparse-domain transfer by testing neighborhoods around sparse seeds rather than isolated samples. Collectively, these results show that the apparent best model is conditional on the specific OOD shift considered. Among the evaluated settings, extrapolation and LOCO generally impose the greatest challenge across models.

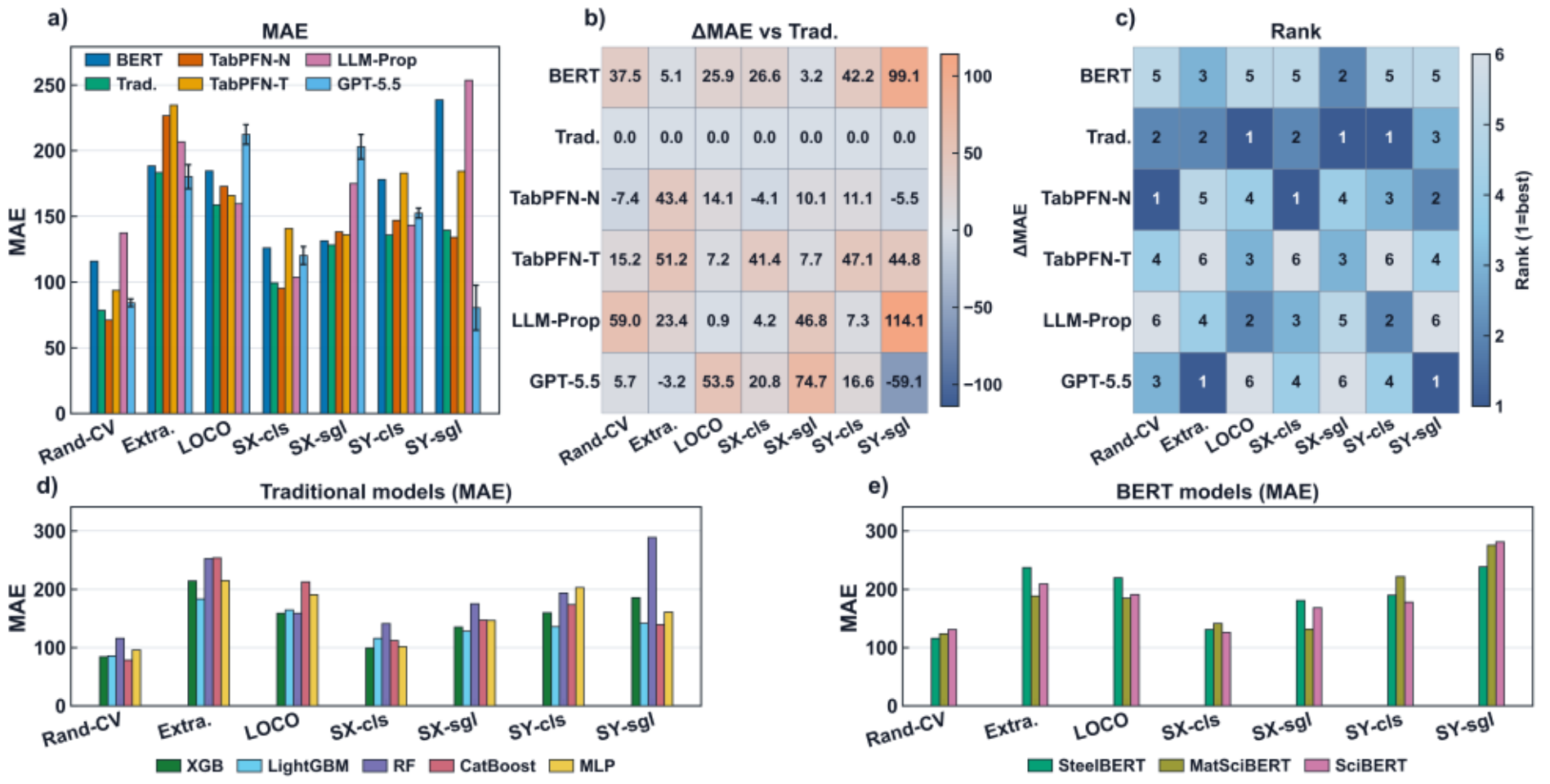

**Figure 6** Model performance comparison for Ti-alloy UTS under Rand-CV and OOD evaluation using full training dataset. a, MAE of six model families, BERT-based, traditional machine-learning models, TabPFN with tabular processing features (TabPFN-N), TabPFN with text-converted processing steps (TabPFN-T), LLM-Prop and GPT-5.5, across Rand-CV and six OOD splits. b, Percentage change in MAE

relative to the traditional-model baseline for each split; positive values indicate higher error than Trad., whereas negative values indicate lower error. c, Within-split model ranking based on MAE, where rank 1 denotes best-performing model and rank 5 denotes the worst. d,e, MAE of individual traditional machine-learning models and BERT variants, respectively, illustrating the within-family performance variation across OOD protocols. Lower MAE indicates better predictive performance. The error bars for GPT-5.5 in the figure represent the standard deviation across three runs.

Table 12,13 summarize the most severe OOD cases after normalizing the Wasserstein distance of each split by the corresponding Rand-CV baseline. Table 12 reports the representation-space severity ratio, $R_x = W_1^*(x) / W_{1,\text{Rand-CV}}^*(x)$, whereas Table 13 reports the target-space severity ratio, $R_y = W_1^*(y) / W_{1,\text{Rand-CV}}^*(y)$. In both cases, larger ratios indicate stronger distribution shift relative to the in-distribution Rand-CV setting. Under representation-space-ratio-extreme splits, GPT-5.5 is best in four of eight tasks, traditional models are best for HEA-UTS, HEA-El, Steel-El and Matbench Steel-YS, and BERT is best for Steel-UTS and Ti-El. Under target-space-ratio-extreme splits, GPT-5.5 achieves the lowest MAE in seven of eight tasks, with LLM-Prop performing best for Al-UTS. These results show that GPT-5.5 is frequently the most competitive model under severe OOD shift, particularly for extrapolation settings, but they also indicate that no single model is uniformly optimal across all alloy–property tasks and shift definitions.

**Table 11** Average model ranking across OOD evaluation tasks using full training dataset. Average ranks of traditional machine-learning models, BERT, TabPFN-N, TabPFN-T and GPT-5.5 across different alloy property prediction tasks. Lower values indicate better relative performance within each task. The table summarizes task-dependent model competitiveness under OOD evaluation, showing that no single model uniformly dominates across all alloy systems and target properties.

| Task | Trad. | BERT | TabPFN-N | TabPFN-T | LLM-Prop | GPT-5.5 |
|---|---|---|---|---|---|---|
| Al-UTS | 3.33 | 5.17 | **2.17** | 2.50 | 4.67 | 3.17 |
| Ti-UTS | **1.67** | 4.17 | 3.17 | 4.67 | 3.67 | 3.67 |
| HEA-UTS | 2.50 | 3.50 | 2.33 | 5.00 | 5.83 | **1.83** |
| Steel-UTS | 3.00 | 3.33 | 4.17 | 3.33 | 5.67 | **1.50** |
| Ti-El | 3.17 | 4.00 | 2.83 | 3.83 | 5.00 | **2.17** |
| HEA-El | **1.83** | 3.33 | 3.17 | 4.33 | 5.67 | 2.67 |
| Steel-El | 3.00 | 4.50 | 4.17 | 3.00 | 5.33 | **1.00** |
| Matbench Steel-YS | 3.50 | **2.33** | 3.33 | 3.33 | 6.00 | 2.50 |

**Table 12** Representation-space OOD severity ratio and model performance across alloy-property tasks using full training dataset. For each task, the most severe representation-space OOD split was identified using the Rand-CV-normalized

Wasserstein ratio, $R_x = W_1^*(x)/W_{1,\text{Rand-CV}}^*(x)$, where $W_1^*(x)$ is the standardized representation-space Wasserstein distance of a given split and $W_{1,\text{Rand-CV}}^*(x)$ is the corresponding value for Random-CV. A larger $R_x$ indicates a stronger representation-space distribution shift relative to the in-distribution baseline. "W ratio" reports $R_x$. "Best" denotes the model with the lowest MAE on the selected split. Delta values indicate the percentage MAE increase of each baseline model relative to the best-performing model; 0% indicates the best or tied-best performance.

| Task | OOD method | $R_x$ | Best | MAE | ΔTrad (%) | ΔBERT (%) | ΔTABPFN-N (%) | ΔTABPFN-T (%) | ΔLLM-Prop | ΔGPT-5.5(%) |
|---|---|---|---|---|---|---|---|---|---|---|
| Al-UTS | SX-sgl | 3.19 | TabPFN-T | 17.20 | 44.19 | 242.0 | 41.17 | 0.00 | 67.47 | 68.83 |
| Ti-UTS | LOCO | 3.09 | Trad. | 158.76 | 0.00 | 16.33 | 8.90 | 4.56 | 0.58 | 33.69 |
| HEA-UTS | LOCO | 7.83 | GPT-5.5 | 195.78 | 3.63 | 47.54 | 3.58 | 14.06 | 51.45 | 0.00 |
| Steel-UTS | SX-cls | 5.84 | BERT | 175.86 | 15.05 | 0.00 | 27.06 | 15.26 | 26.52 | 20.17 |
| Ti-El | LOCO | 3.09 | BERT | 7.30 | 9.76 | 0.00 | 6.84 | 12.51 | 8.44 | 15.37 |
| HEA-El | LOCO | 7.83 | GPT-5.5 | 13.46 | 6.49 | 31.00 | 9.39 | 17.21 | 23.77 | 0.00 |
| Steel-El | SX-cls | 5.84 | GPT-5.5 | 7.91 | 30.77 | 59.50 | 43.45 | 42.92 | 36.77 | 0.00 |
| Matbench Steel-YS | SX-sgl | 5.65 | GPT-5.5 | 67.91 | 71.18 | 18.67 | 64.12 | 29.44 | 320.46 | 0.00 |

**Table 13** Target-space OOD severity ratio and model performance across alloy-property tasks using full training dataset. For each task, the most severe target-space OOD split was identified using the Rand-CV-normalized Wasserstein ratio, $R_y = W_1^*(y)/W_{1,\text{Rand-CV}}^*(y)$, where $W_1^*(y)$ is the standardized target-space Wasserstein distance of a given split and $W_{1,\text{Rand-CV}}^*(y)$ is the corresponding value for Random-CV. A larger Ry indicates a stronger target-space distribution shift relative to the in-distribution baseline. "W ratio" reports Ry. "Best" denotes the model with the lowest MAE on the selected split. Delta values indicate the percentage MAE increase of each baseline model relative to the best-performing model; 0% indicates the best or tied-best performance.

| Task | OOD method | $R_y$ | Best | MAE | ΔTrad (%) | ΔBERT (%) | ΔTABPFN-N (%) | ΔTABPFN-T (%) | ΔLLM-Prop | ΔGPT-5.5 (%) |
|---|---|---|---|---|---|---|---|---|---|---|
| Al-UTS | Extra. | 16.85 | 6.38 | LLM- | 84.26 | 7.81 | 8.32 | 15.11 | 16.52 | 0.00 |
| | | | | Prop | | | | | | |
| Ti-UTS | Extra. | 18.16 | 5.49 | GPT-5.5 | 180.16 | 1.78 | 4.59 | 25.85 | 30.22 | 14.74 |
| HEA-UTS | Extra. | 21.18 | 3.47 | GPT-5.5 | 316.61 | 5.96 | 27.27 | 20.94 | 36.90 | 59.10 |
| Steel-UTS | Extra. | 31.93 | 2.23 | GPT-5.5 | 382.63 | 53.31 | 52.37 | 55.82 | 50.14 | 54.94 |
| Ti-El | Extra. | 11.65 | 2.45 | GPT-5.5 | 12.34 | 10.10 | 6.88 | 5.82 | 5.13 | 43.23 |
| HEA-El | Extra. | 14.45 | 3.62 | GPT-5.5 | 22.76 | 14.73 | 22.30 | 16.70 | 29.54 | 54.71 |
| Steel-El | Extra. | 40.55 | 6.08 | GPT-5.5 | 24.50 | 40.41 | 40.73 | 39.56 | 38.46 | 40.86 |
| Matbench Steel-YS | Extra. | 28.34 | 4.11 | GPT-5.5 | 369.46 | 9.12 | 2.03 | 14.46 | 16.36 | 51.84 |

## Error against OOD severity: a quantitative relation

The severity metric has so far functioned as a label rather than as a predictor. Here the error-versus-severity relation is examined as an actionable quantity in its own right, using a complete joint analysis of a fold-level severity dataset that reports both $R_x$ and $R_y$ for every split across all 7 protocols (RandCV, LOCO, Extra., SX-cls, SY-cls, SX-sgl, SY-sgl). For 8 tasks — Al-UTS, Ti-UTS, HEA-UTS, Steel-UTS, Ti-El, HEA-El, Steel-El and Matbench Steel-YS — and 6 model families (Traditional, BERT, TabPFN-N, TabPFN-T, LLM-Prop, GPT-5.5), each protocol split supplies the MAE, the matched Rand-CV baseline MAE, and the two severity ratios at the same split.

With D=MAE(split)/MAE(Rand-CV) and both severities entering as base-2 logarithms, each model family is fitted by ordinary least squares as

$$log_2 D = a(task) + b_x log_2 R_x + b_y log_2 R_y + \varepsilon \quad (3)$$

where a (task) is a full set of task fixed effects (one dummy per task, no separate intercept), so that each task's own baseline difficulty is absorbed and $b_x$ and $b_y$ are identified from how the degradation ratio moves within a task as its protocol, and hence its severity, changes. This gives 56 points per model family (8 tasks × 7 protocols). Intervals are the standard OLS slope intervals with a Student-t critical value at 46 degrees of freedom (8 task dummies plus 2 slopes; $t_{0.975,46}$=2.013), and $R^2$ is the fitted-model coefficient of determination with task effects included. The result is summarized in Table 14.

**Table 14** Summary of the joint error-versus-severity regression fit using the full

training dataset. All models are estimated incorporating $log_2D = a(task) + b_x log_2 R_x + b_y log_2 R_y + \varepsilon$, task fixed effects, n=56 per model. The final row is the mean-predictor reference fit to its own degradation curve under the identical specification and is reported for the reliability analysis, not as a model family.

| Model | n | $b_x$ | 95% CI ($b_x$) | $b_y$ | 95% CI ($b_y$) | $R^2$ |
|---|---|---|---|---|---|---|
| **Traditional** | 56 | 0.049 | [-0.056, 0.153] | 0.365 | [0.311, 0.418] | 0.825 |
| **BERT** | 56 | 0.021 | [-0.104, 0.145] | 0.325 | [0.261, 0.388] | 0.748 |
| **TabPFN-N** | 56 | 0.054 | [-0.061, 0.169] | 0.370 | [0.311, 0.429] | 0.801 |
| **TabPFN-T** | 56 | 0.029 | [-0.117, 0.175] | 0.352 | [0.278, 0.427] | 0.719 |
| **LLM-Prop** | 56 | -0.087 | [-0.189, 0.014] | 0.356 | [0.304, 0.408] | 0.846 |
| **GPT-5.5** | 56 | 0.051 | [-0.107, 0.209] | 0.265 | [0.184, 0.346] | 0.532 |
| **Mean-predictor baseline** | 56 | -0.238 | [-0.334, -0.142] | 0.370 | [0.321, 0.420] | 0.846 |

Once $R_y$ enters the model, $b_x$ is not distinguishable from zero for any of the 6 model families: every 95% confidence interval spans zero, with point estimates in the range -0.087–0.054. $b_y$, by contrast, remains clearly positive, in the range 0.265–0.370, and every interval excludes zero. Target-space severity, rather than representation-space severity, is therefore the reliable predictor of error once the two are modelled together. The weakest target-space slope belongs to GPT-5.5 ($b_y$=0.265, $R^2$=0.532), which degrades slowly in absolute terms but is also the least completely described by the log-linear specification. This is a statement about how much of GPT-5.5's error variance severity accounts for, not about its absolute accuracy, and should not be read as GPT-5.5 performing worse under shift than the other families: Table 14 and Figure 7 describe the shape of the relationship, not the level of error.

The two severity axes are only mildly correlated across the 56 task-by-protocol cells (Pearson r=0.251 overall, 0.182 within task), and the variance inflation factor of both slopes is close to one (1.25 for $b_x$, 1.14 for $b_y$), well under the conventional concern threshold of five, so the null result for $b_x$ is not an artefact of the two severities being confounded. Figure 7, 8 show the corresponding partial-regression panels: each panel plots the residual of $\log_2 D$ against the residual of one severity axis after removing task effects and the other axis, so by the Frisch-Waugh-Lovell theorem the drawn slope equals the coefficient in Table 14 exactly. No panel of Figure 7 has an interval excluding zero, whereas every panel of Figure 8 does.

Two further checks support the result. First, the uncoupled single-axis fits ($\log_2 D \sim a(task) + \log_2 R_x$ alone) give representation-space slopes in the range 0.039–0.186, larger than the joint estimates but still individually insignificant, so the joint specification does not manufacture the null result. Second, refitting on the 80 individual

Random-CV and LOCO folds of each family, rather than on their protocol-level aggregate, preserves the qualitative picture: $b_y$ stays positive and significant for every family (0.224–0.280) and $b_x$ stays in the range 0.027–0.131, while TabPFN-N shows a marginal positive representation-space interval. Fold-level estimates are reported as point comparisons only, since folds within a protocol are not independent draws and a properly clustered standard error was not computed for this check.

The reliability-threshold question — the severity at which a model's fitted error reaches a mean-predictor baseline that always returns the training-split mean — is answered with the same fold-level severity data and the same task-fixed-effects specification, fitted to the baseline's own degradation curve ($b_x$=-0.238, 95% CI [-0.334, -0.142]; $b_y$=0.370, 95% CI [0.321, 0.420]). Across the full severity range observed in this study ($R_y$ up to approximately 41), no model family degrades faster than the reference predictor, so the fitted curves cannot cross at any severity: 0 of 48 task-level curve pairs cross inside the observed range, so no family is overtaken by the baseline. Every model's Rand-CV error is already 1.1–3.8× lower than the baseline's, no model's target-space slope exceeds the baseline's (closest: TabPFN-N, 0.0004 below it), and even at the highest observed severity the smallest fitted advantage is 1.11× (LLM-Prop), so the lead over the mean predictor is preserved or widened throughout. Solving the fitted curves for an algebraic crossing point returns severities orders of magnitude beyond the observed range, which is not reported as a threshold because a log-linear fit extrapolated that far has no claim to validity. One secondary observation: the baseline's own representation-space slope is negative with an interval excluding zero, unlike the model families, whose representation-space slopes are indistinguishable from zero; this most likely reflects which specific LOCO clusters happen to be held out rather than a general property of the mean-predictor baseline, and we note it without building further claims on it. These findings are consistent with the LLM results obtained using the optimized k-NN configuration for reference retrieval, as shown in Fig. S50. This agreement suggests that selecting reference samples using the k-NN-optimal k and the corresponding representation space can yield conclusions similar to those obtained using the entire training set as reference context. This finding is particularly important for evaluating LLM performance under OOD conditions, as it supports replacing full training-set context with a compact set of relevant reference samples, thereby reducing input size and inference costs while preserving the main conclusions.

GPT-5.5's three repeats per fold let this inference uncertainty be quantified directly, alongside the sampling uncertainty Table 14 already reports. The five baseline model families are deterministic given a fixed seed and have no corresponding repeat structure, so this treatment applies to GPT-5.5 only. Refitting the same joint regression separately on each of the three runs — using that run's own MAE at every fold, including its own Rand-CV baseline, rather than the 3-run mean used in Table 14 — gives three independent estimates: $b_x$ = 0.041, 0.109 and −0.020; $b_y$ = 0.238, 0.257 and 0.299.

The spread across these three estimates (SD = 0.064 for $b_x$, 0.031 for $b_y$) is a direct, empirical measure of how much GPT-5.5's output stochasticity alone moves the slope estimates, independent of the between-task, between-protocol sampling uncertainty the original confidence intervals already capture. Combining the two sources as independent variances (total SE² = sampling SE² + run-to-run SD²) widens GPT-5.5's intervals to $b_x$ = 0.051, 95% CI [−0.153, 0.255], and $b_y$ = 0.265, 95% CI [0.162, 0.368] — roughly 30% wider than the sampling-only intervals in Table 14, but neither conclusion changes: $b_x$ remains indistinguishable from zero and $b_y$ remains clearly positive and significant even once inference uncertainty is added on top of sampling uncertainty.

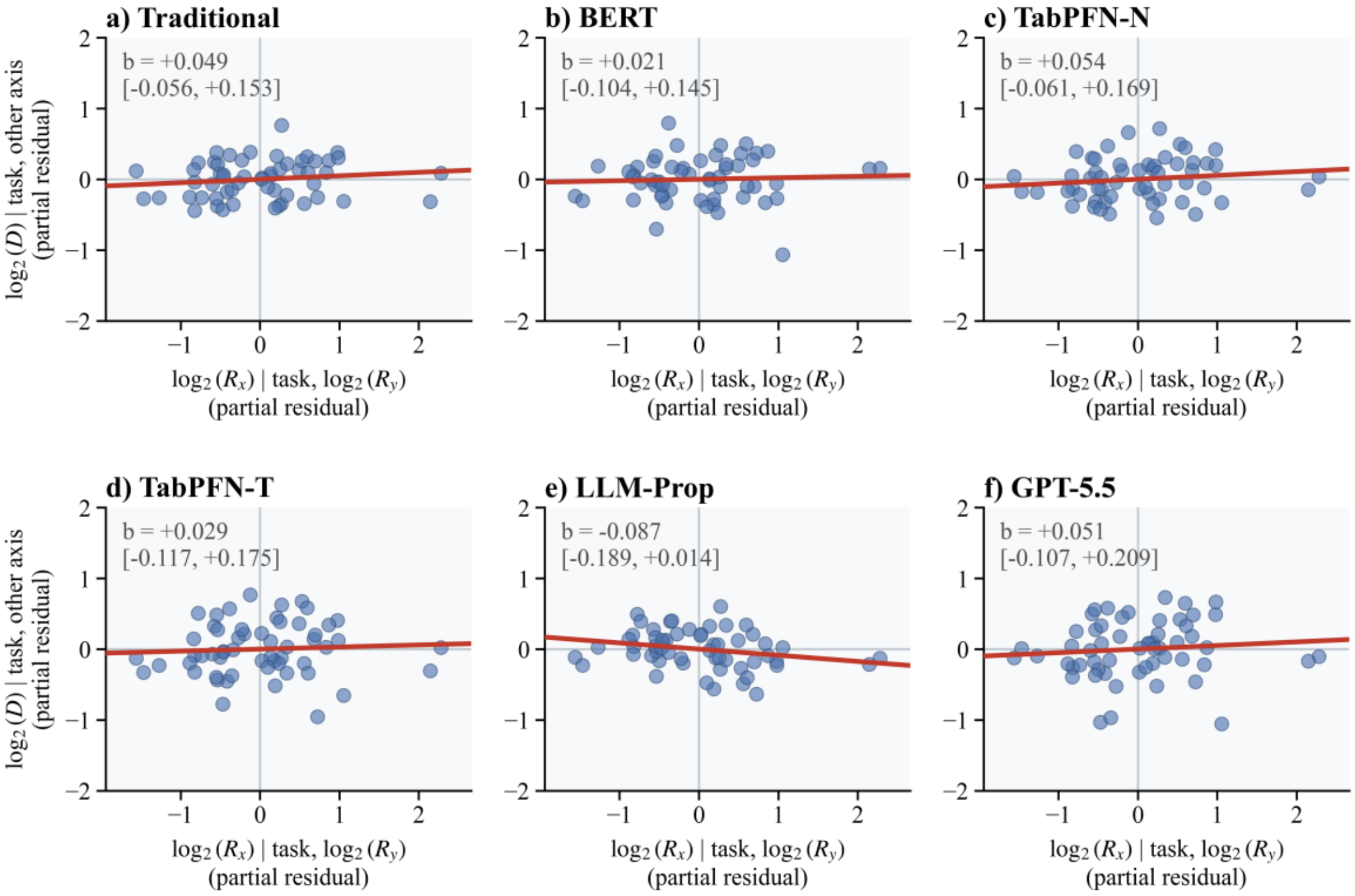

**Figure 7** Representation-space partial-regression analysis across six model families using full training dataset. Panels (a–f) show the partial relationship between representation-space severity and prediction degradation after removing the effects of task identity and target-space severity. According to the Frisch–Waugh–Lovell theorem, the slope of each fitted line is identical to the corresponding coefficient $b_x$ from the full multivariable regression. Red lines denote the fitted partial-regression trends, with the estimated slope and 95% confidence interval reported in each panel. All confidence intervals include zero, indicating no statistically significant independent association between representation-space severity and prediction degradation.

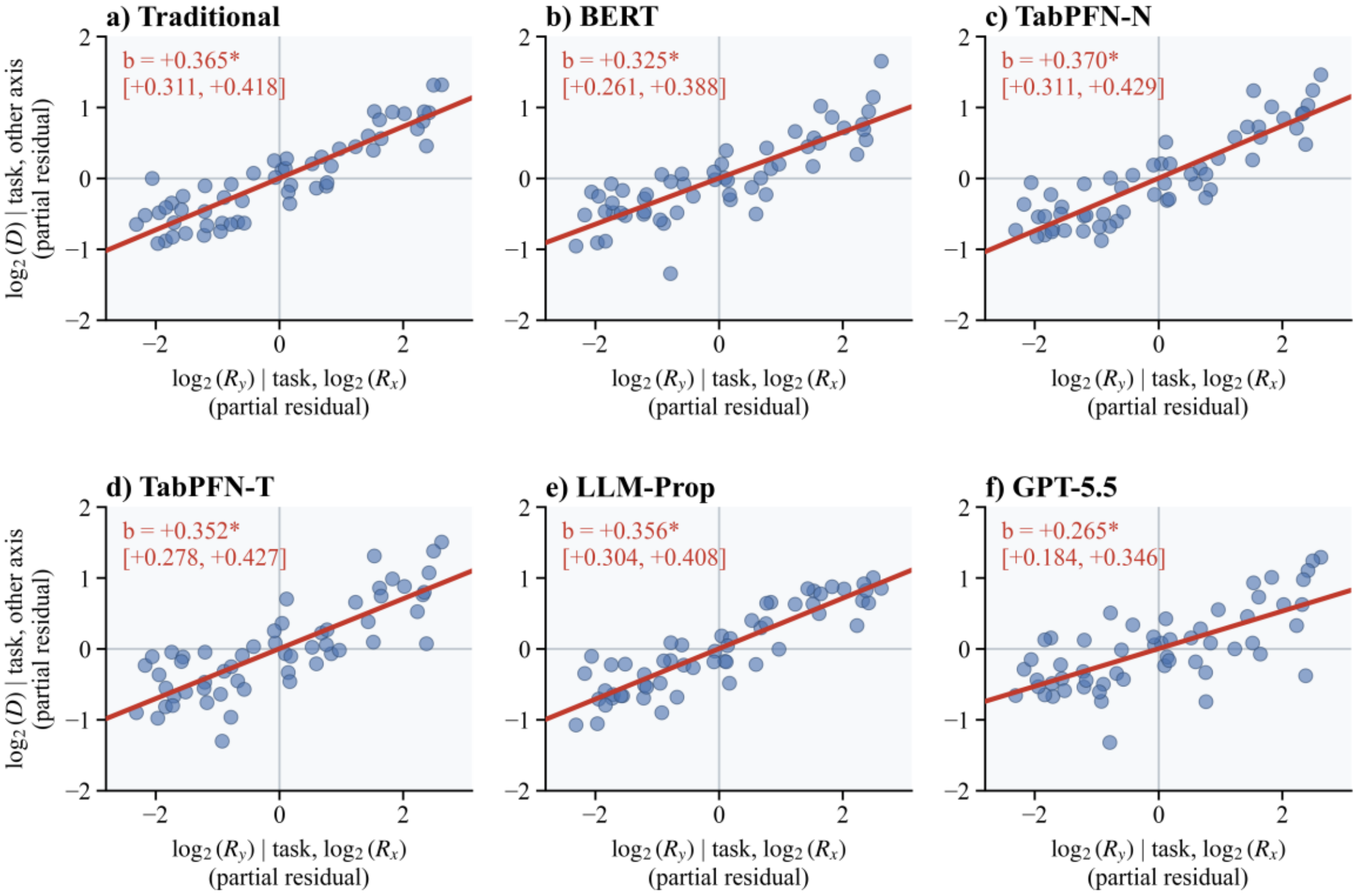

**Figure 8** Target-space partial-regression analysis across six model families using the full training dataset. Panels (a–f) show the partial relationship between target-space severity and prediction degradation after removing the effects of task identity and representation-space severity. Panels (a–f) show the partial relationship between representation-space severity and prediction degradation after removing the effects of task identity and target-space severity. According to the Frisch–Waugh–Lovell theorem, the slope of each fitted line is identical to the corresponding coefficient $b_y$ from the full multivariable regression. Red lines denote the fitted partial-regression trends, with the estimated slope and 95% confidence interval reported in each panel. All slopes are positive and their confidence intervals exclude zero, indicating a statistically significant independent association between target-space severity and prediction degradation.

**Optimal k-NN references and LLM-reported confidence**

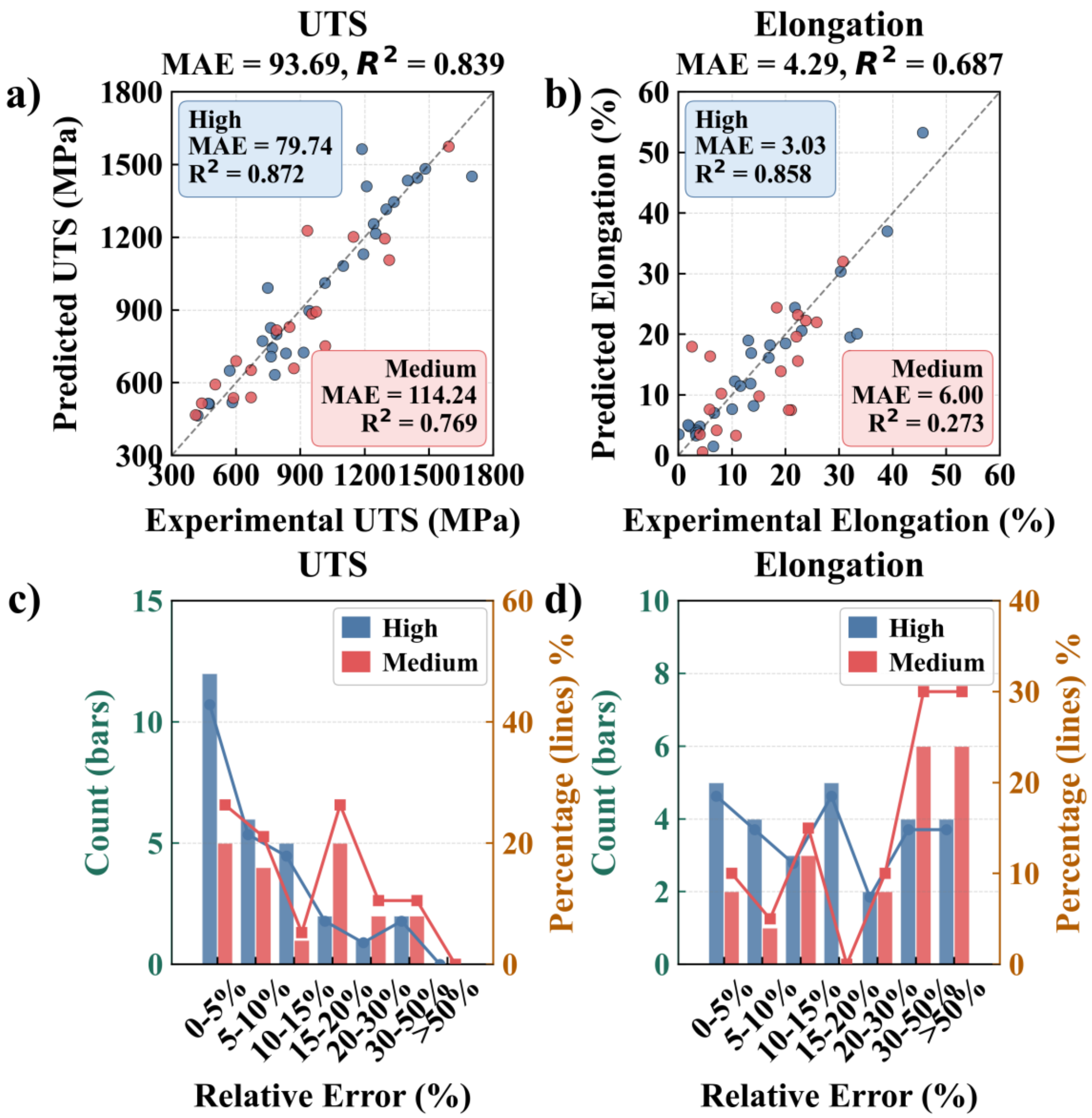

**Figure 9** Prediction accuracy and error distribution of Gemini-2.5-Flash for titanium alloy property prediction using full training dataset. Scatter plots show predicted versus experimental values for multi-target prediction of (a) UTS and (b) elongation, and single-target prediction of (c) UTS and (d) elongation. Samples are grouped according to the confidence levels reported by the LLM (High: blue; Medium: red). The corresponding MAE and $R^2$ values are shown for each confidence category. Relative error distributions are presented in panels (c) and (d). The bars indicate the number of samples (left axis), whereas the lines represent the corresponding percentages (right axis).

LLM-reported confidence is generally associated with prediction accuracy across the examined alloy systems and prediction settings, with higher-confidence predictions exhibiting lower MAE in most cases, as shown in Figure 9 and Figs. S28-34. For titanium alloy prediction using Gemini-2.5-Flash, high-confidence UTS predictions achieve an MAE of 79.74 MPa, compared with 114.24 MPa for medium-confidence predictions, with corresponding $R^2$ values of 0.872 and 0.769, respectively. A similar relationship is observed for elongation, where high-confidence predictions yield a lower MAE of 3.03 percentage points and a higher $R^2$ of 0.858 than medium-confidence predictions, whose corresponding values are 6.00 percentage points and 0.273. The relative-error distributions in Figure 9 c–d further support this relationship. For both properties, a larger proportion of high-confidence predictions falls within the 0–5% relative-error interval, whereas medium-confidence elongation predictions exhibit a greater proportion of errors exceeding 30%. Together, these results demonstrate that the confidence labels contain useful information about prediction reliability, with higher confidence generally indicating a greater likelihood of obtaining a low-error prediction.

The cross-model comparisons in Figs. S31-34 further show that Gemini-3-Flash achieves higher prediction accuracy than Gemini-2.5-Flash. These results suggest that greater model capability can improve alloy property prediction, although improvements in overall accuracy do not necessarily establish better alignment between confidence and prediction error. The observed confidence-dependent error patterns nevertheless support the use of LLM-reported confidence as a diagnostic indicator for identifying potentially less reliable predictions. Because higher confidence does not guarantee lower error in every task or prediction setting, the subsequent analysis systematically quantifies the ordering of MAE across confidence levels under seven data-splitting protocols. This evaluation assesses how consistently the confidence–error relationship holds under random splitting and different forms of distribution shift, providing a broader basis for evaluating the reliability of self-reported confidence in OOD prediction.

To quantitatively evaluate the monotonic relationship between confidence and prediction error, we calculated the fraction of task–protocol units satisfying the confidence ladder relationship: $MAE_{high} < MAE_{medium} < MAE_{low}$. As shown in Figure 10, protocols exhibit the expected monotonic confidence–error trend. Random-CV achieves the highest consistency ratio (87.5%), whereas all other evaluation protocols maintain at least 50% agreement with this relationship, indicating that LLM-reported confidence can effectively reflect prediction uncertainty under diverse distribution shifts.

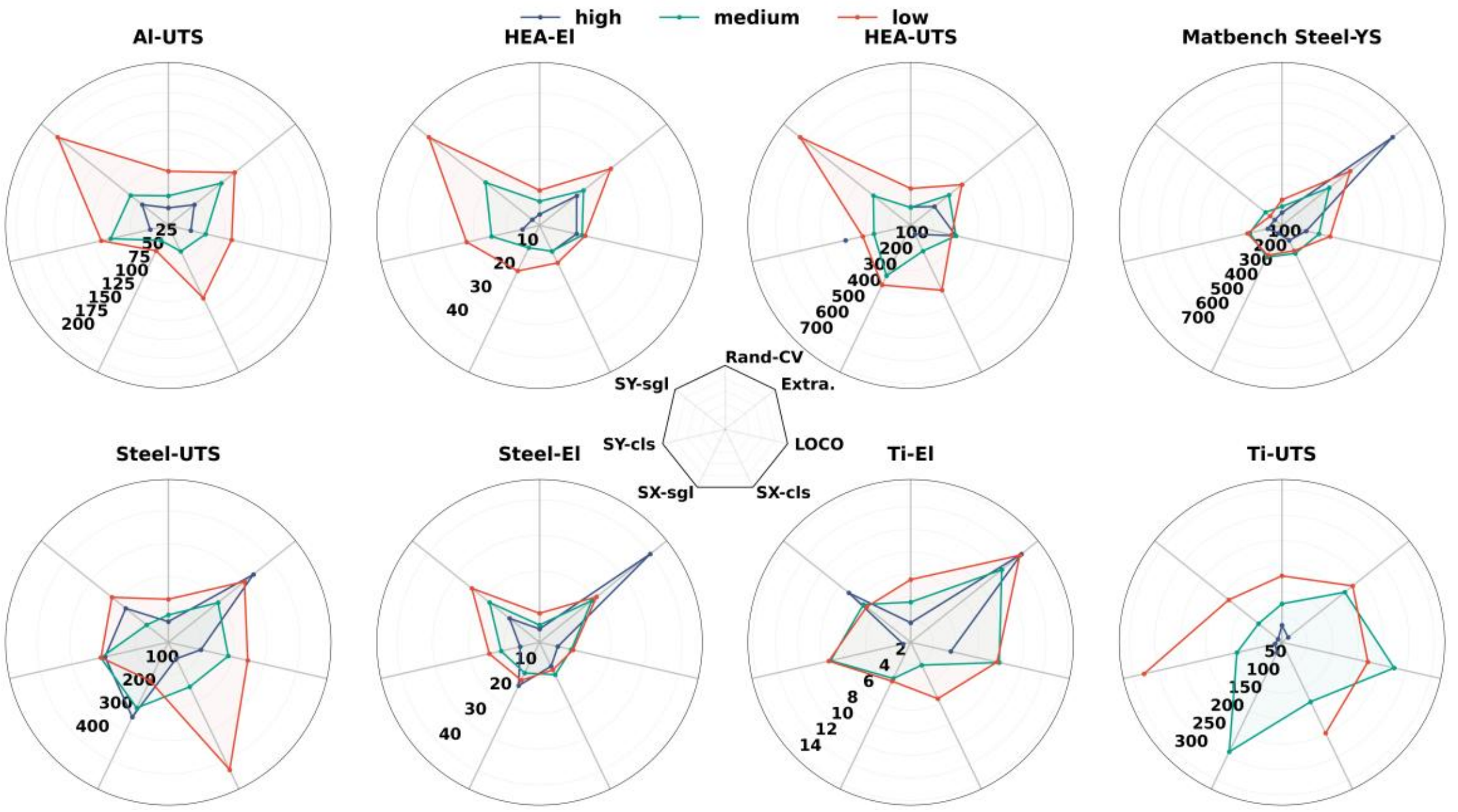

**Figure 10** Confidence–error radar charts across prediction tasks using the k-NN-optimal k reference setting. Each panel shows the mean absolute error (MAE) of LLM predictions under seven evaluation protocols for one alloy property prediction task. The angular axes represent Rand-CV, Extra., LOCO, SX-cls, SX-sgl, SY-cls, and SY-sgl, with their full names and ordering indicated by the central reference radar. The three curves correspond to LLM self-reported confidence levels of high, medium, and low. Radial values represent MAE in the original target units, with MPa for UTS tasks and % for elongation tasks. Task-specific radial scales are used; therefore, MAE values are directly comparable only within individual panels.

## Effect of OOD severity on LLM-reported confidence

Finally, we examine the effect of OOD severity on LLM-reported confidence and prediction error from GPT-5.5 for the Titanium UTS dataset. Figure 11a–b illustrates how titanium alloy UTS prediction changes with increasing OOD severity measured in representation space ($R_x$). Compared with Rand-CV, OOD splits generally exhibit a reduced proportion of high-confidence predictions and increased fractions of medium- and low-confidence predictions. The fraction of high-confidence predictions decreases with increasing $R_x$, whereas low-confidence predictions become more prevalent under larger representation-space shifts (Figure 11b). Medium-confidence predictions serve as an intermediate regime with a more variable distribution across different splits. Similar trends are observed in the other alloys (Figs. S51,53), while the corresponding ($R_y$)-based analyses show qualitatively similar but less stable behavior (Figs. S52, 54).

This suggests that LLM-reported confidence is sensitive to the degree of distribution shift in representation space, although it should not be interpreted as a calibrated uncertainty estimate.

Figure 11c and Figs. S55,57 further examine the relationship between representation-space shift and prediction error. Unlike the confidence redistribution observed in Figure 11a–b, the mean relative error does not exhibit a clear monotonic increase with increasing $R_x$. The considerable variation amongst different OOD splits indicates that larger representation-space distances do not necessarily lead to proportionally larger prediction errors, which is consistent with the results shown in Figure 7. Nevertheless, the aggregate comparison Figure 11 d demonstrates that OOD evaluation generally results in higher prediction errors than Rand-CV, with the $R_x$ subsets showing further degradation. Combined with the confidence-dependent error analysis in Figs. S55-58, these results indicate that LLM-reported confidence provides complementary reliability information under distribution shift: confidence responds more consistently to representation-space novelty than the absolute prediction error, making it a useful indicator for identifying potentially unreliable predictions in OOD materials screening.

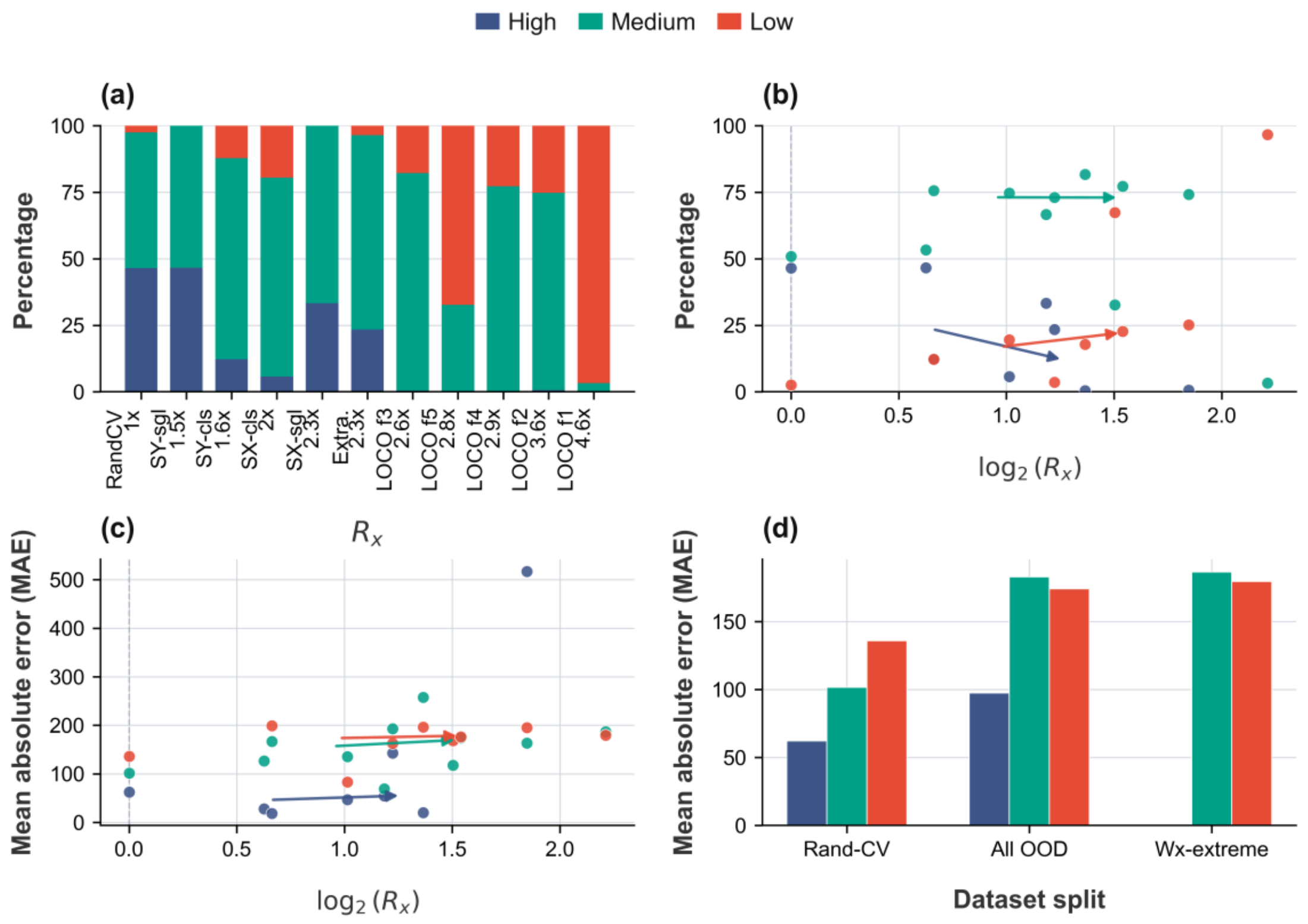

**Figure 11** Effect of OOD severity on LLM-reported confidence and prediction error of GPT-5.5 for the Titanium UTS dataset using full training dataset. a, Proportions of high-, medium- and low-confidence predictions under different dataset splits, with OOD severity ($R_X$). LOCO f1–5 denote five leave-one-cluster-out splits, where f1–f5 correspond to five predefined cluster-level OOD test subsets generated by excluding one cluster at a time from the training data. b, Confidence-level proportions as a function of OOD severity, expressed as $\log_2(R_X)$. c, Mean relative error for each confidence level as a function of OOD severity. Arrows in b and c indicate the sign of the Spearman rank correlation between $log_2(R_x)$ and the plotted quantity, calculated separately for each confidence level using all points with nonzero confidence share, with positive, negative and zero correlations represented by upward-right, downward-right and horizontal-right arrows, respectively. d, Aggregated mean relative error for Rand-CV, all OOD splits and $R_x$ splits. Increasing OOD severity generally reduces the fraction of high-confidence predictions and increases prediction error, while high-confidence predictions remain associated with lower relative-error outcomes.

**Comparison of LLM confidence with retrieval density**

To test whether the confidence–error relationship in Figure 11 reflects something beyond retrieval geometry, we compared LLM-reported confidence with density-based controls built from the same retrieved reference samples. Density was computed from cosine similarities in the standardized numerical feature space, then binned into High, Medium and Low categories matching the LLM's own class proportions (Methods **Density-based control construction**).

As shown in Table 15, LLM-reported confidence tracked error more consistently than density-based confidence under both reference settings: 37/56 task–protocol units (66.1%) satisfied the expected confidence ladder for LLM-reported confidence versus 33/56 (58.9%) for density, using k-NN-optimal-k references, and 38/56 (67.9%) versus 28/56 (50.0%) using the full training set. A further control — density computed from the k-NN-optimal-k references but binned to match LLM@full-k's class proportions — reached only 26/56 units (46.4%), confirming that the confidence–error ordering is not simply a byproduct of retrieval density. Expanding the reference set from k-NN-optimal-k to the full training set slightly improved LLM consistency (66.1% → 67.9%) but reduced density-based consistency (58.9% → 50.0%). This divergence suggests additional references give the LLM broader contextual evidence for judging reliability, rather than simply increasing local similarity to the target.

We further compared prediction errors within matched confidence bins between LLM-reported confidence and density-based controls (Table 16). Because an informative low-confidence bin should carry a larger, not smaller, error, the comparison applies the asymmetric criterion described in Methods **Density-based control construction**. By this measure, LLM-reported confidence gave the slightly better-discriminating bin in seven of nine comparisons. The two exceptions all favored density-based confidence: the high-confidence bin under the optimal-k setting (34.45 versus 31.9), and the high-confidence bin under the full-k setting (49.45 versus 48.35). The small differences in MAE are likely due to means over tasks, protocols and different types of datasets; for example, UTS has large values compared to elongation, which is far noisier due to measurement issues. However, if we consider LLM < Density and Density < LLM of paired task–protocol units applying the same asymmetric criterion for MAE as before, then the conclusion (eight of nine) remains unchanged.

Together, these results show that LLM-reported confidence carries reliability information beyond simple retrieval similarity. Because this consistency survives even with the compact k-NN-optimal-k reference set, a small, well-chosen reference set can match full-reference reliability assessment at lower context length and inference cost.

**Table 15** Statistical monotonic confidence ladder comparison between LLM-reported confidence and density-based controls. Values represent the number and fraction of task–protocol units satisfying the expected monotonic relationship between confidence level and prediction error, with percentages shown in parentheses. LLM@optimal-k and LLM@full-k denote confidence reported by the LLM using k-NN-optimal k and full-reference settings, respectively. Denstiy@optimal-k and Density@full-k denote density-based controls calculated from the corresponding retrieved reference sets. Density@optimal-k (full-k bins) uses density calculated from optimal-k reference samples, while assigning High/Medium/Low bins according to the class proportions generated by LLM@full-k, thereby isolating the influence of LLM confidence assignment from density differences.

| **Method** | **Density@optimal-k** | **LLM@optimal-k** | **Density@optimal-k (full-k bins)** | **Density@full-k** | **LLM@full-k** |
| --- | --- | --- | --- | --- | --- |
| **Random-CV** | **8/8 (100.0%)** | **7/8 (87.5%)** | **7/8 (87.5%)** | **6/8 (75.0%)** | **8/8 (100.0%)** |
| **Extra.** | **5/8 (62.5%)** | **4/8 (50.0%)** | **3/8 (37.5%)** | **5/8 (62.5%)** | **5/8 (62.5%)** |
| **LOCO** | **6/8 (75.0%)** | **5/8 (62.5%)** | **4/8 (50.0%)** | **4/8 (50.0%)** | **5/8 (62.5%)** |
| **SX-cls** | **3/8 (37.5%)** | **5/8 (62.5%)** | **2/8 (25.0%)** | **3/8 (37.5%)** | **5/8 (62.5%)** |
| **SX-sgl** | **4/8 (50.0%)** | **5/8 (62.5%)** | **3/8 (37.5%)** | **3/8 (37.5%)** | **5/8 (62.5%)** |
| **SY-cls** | **4/8 (50.0%)** | **6/8 (75.0%)** | **5/8 (62.5%)** | **4/8 (50.0%)** | **7/8 (87.5%)** |
| **SY-sgl** | **3/8 (37.5%)** | **5/8 (62.5%)** | **2/8 (25.0%)** | **3/8 (37.5%)** | **3/8 (37.5%)** |
| **ALL** | **33/56 (58.9%)** | **37/56 (66.1%)** | **26/56 (46.4%)** | **28/56 (50.0%)** | **38/56 (67.9%)** |

**Table 16** Confidence-bin prediction error comparison between LLM-reported confidence and density-based controls. LLM and Density denote LLM-reported confidence and density-based confidence, respectively. MAE comparisons are performed within matched task–protocol units for each confidence category. Density-based controls are generated from retrieval similarity using the corresponding reference samples, with High/Medium/Low bins assigned according to the same class proportions as the LLM confidence categories. For the full-k comparison, density calculated from optimal-k references is additionally matched to the confidence-bin proportions generated by LLM@full-k. LLM < Density and Density < LLM indicate the numbers of paired task–protocol units where the corresponding confidence assignment achieves lower MAE. Because an informative low-confidence bin should flag the least reliable predictions rather than accurate ones, Winner uses an asymmetric rule: for the high- and medium-confidence bins, the method with the lower MAE wins; for the low-confidence bin, the method with the higher MAE wins.

| Pair | Confidence | Total units | Paired units | LLM < Density | Density < LLM | Tie | MAE-LLM | MAE-Density | Winner |
|---|---|---|---|---|---|---|---|---|---|
| Density@optimal-k vs LLM@optimal-k | high | 56 | 47 | 13 | 28 | 6 | 34.45 | 31.9 | Density |
| Density@optimal-k vs LLM@optimal-k | medium | 56 | 56 | 30 | 25 | 1 | 71.1 | 71.81 | LLM |
| Density@optimal-k vs LLM@optimal-k | low | 56 | 55 | 21 | 31 | 3 | 116.51 | 116.35 | LLM |
| Density@full-k vs LLM@full-k | high | 56 | 54 | 35 | 17 | 2 | 49.45 | 50.34 | LLM |
| Density@full-k vs LLM@full-k | medium | 56 | 56 | 32 | 24 | 0 | 99.83 | 102.68 | LLM |
| Density@full-k vs LLM@full-k | low | 56 | 47 | 17 | 28 | 2 | 101.73 | 89.49 | LLM |
| Density@optimal-k (LLM full-k bins) vs LLM@full-k | high | 56 | 54 | 29 | 21 | 4 | 49.45 | 48.35 | Density |
| Density@optimal-k (full-k bins) vs LLM@full-k | medium | 56 | 56 | 32 | 22 | 2 | 99.83 | 103.12 | LLM |
| Densityz@optimal-k (full-k bins) vs LLM@full-k | low | 56 | 47 | 15 | 27 | 5 | 101.73 | 90.8 | LLM |

## Discussion

Despite advances, limitations exist. LLMs exhibit inherent stochasticity even at zero temperature, stemming from architectural factors which can introduce variability in outputs. This variability arises from non-deterministic computations during inference, including floating-point precision, parallel execution differences, and sensitivity to token-level perturbations. Here we have fixed the temperature to zero, however, we expect the general trend to be similar as our MAE and $R^2$ are close to each other when the temperature is set to non-zero. Additionally, performance dips in heterogeneous datasets (e.g., steels with inconsistent processing descriptions), highlights the need for standardized data formats. While our self-reported confidence calibration correlates well with accuracy, it relies on prompt engineering, which may not generalize across all models or tasks. Furthermore, the effectiveness of the Guided Mode (which is not used for the OOD analysis) depends on the quality of prior model estimates and the consistency of retrieved reference samples, suggesting that future improvements should focus on better integration of structured data and multimodal information. A further limitation concerns model versioning: because commercial LLM endpoints are updated or withdrawn on a timescale shorter than a typical review cycle, different experiments in this study necessarily draw on different model generations (Table 2), and the Gemini-3-Flash Preview endpoint used for the Matbench comparison was deprecated by the vendor during the course of this work, so those specific results cannot be reproduced against the same endpoint going forward. This is an unavoidable consequence of benchmarking actively-developed commercial systems rather than a methodological choice, but it means the paper's conclusions should be read as holding for the tested model generations specifically, not as a permanent characterization of any named product.

Our results indicate that LLM-reported confidence is an error-informative reliability signal: high-confidence predictions are more likely to fall within low relative-error ranges, whereas medium-confidence predictions are associated with larger and more dispersed errors. This confidence–error relationship provides a practical basis for materials screening, where high-confidence predictions can be prioritized for candidate selection, while medium-confidence cases may require additional validation. A density-only control built from the same retrieved samples, with no LLM reasoning involved, tracks this relationship less reliably — satisfying the expected confidence ordering in 50.0–58.9% of task–protocol units against the LLM's 66.1–67.9% (Tables 15,16) — indicating that self-reported confidence carries reliability information beyond simple retrieval proximity.

Most importantly, we have demonstrated how the Wasserstein distance provides a geometric means to quantify the severity of deviations of OOD from in-distribution data. OOD data from extrapolation and cluster analysis show, in general, the greatest severity and deviations from Random-CV results. Previous studies on OOD applied to materials have lacked such a quantitative measure. In terms of models, the LLM GPT-5.5 is often strongest under severe OOD shifts, but the benchmark does not support a universal-outperformer. This competitiveness is not confined to the alloy systems that form the core of this benchmark: on two Matbench tasks spanning a different materials class and representation — composition-based steel yield strength, and crystal-structure-based exfoliation energy encoded as text via Robocrystallographer — GPT-5.5 and Gemini-3-Flash Preview matched or exceeded specialized leaderboard models, including graph-neural-network architectures purpose-built for structural inputs (Table 6). This suggests the approach's competitiveness generalizes beyond the alloy-specific setting studied here, though two tasks are not a substitute for a dedicated cross-domain benchmark. Whether the relatively stronger performance of GPT-5.5 under severe shift reflects genuine reasoning over retrieved references, rather than a more favorable prior or reference set, has not been directly tested. A representative worked example confirms that the reasoning field can reflect physically consistent composition–processing–property relationships (Figs. S2,3), but this is one manually-checked case, not the systematic multi-sample audit that would be needed to test the OOD-severity-specific claim directly; that broader audit has not been performed. LLM-Prop, the fine-tuned transformer model, that has been shown recently to outperform several other models, including chatbots, does not perform particularly well on the relatively small datasets studied here[34]. Not surprisingly, we also find that the fraction of high-confidence predictions from the models is reduced as the OOD severity increases statistical prediction error, while lower relative-error outcomes are associated with high-confidence predictions. Models robust under OOD generalization relative to Random-CV predictions are key to exploration of new materials not represented in the training data, and our work provides a step in this direction.

Direct comparison against k-NN baselines shows that the LLM's advantage is not simply retrieval in disguise: at a fixed 5% reference budget it improves on the best k-NN configuration on eight tasks, and even against a k-NN baseline tuned directly on the OOD test data it achieves lower error in 61.67% of 120 task–protocol configurations, most consistently under LOCO and least consistently under isolated sparse-feature test points. A matched-cohort comparison using a temporally recent sample set found no substantial or target-consistent advantage for the literature-compiled cohort, which argues against memorization as the primary driver of the reported accuracy, though a small or target-specific effect cannot be excluded outright given the cohort size. A joint error-versus-severity analysis (Table 14), fit with task fixed effects across all seven OOD protocols, finds that error increases reliably with target-space severity for every model family tested ($b_y$ = 0.27–0.37, all CIs excluding zero), while representation-space severity shows no reliable independent effect for any model family once target-space severity is accounted for (all $b_x$ CIs span zero) — including for GPT-5.5, the model now used throughout this joint analysis in place of GPT-5.4, whose predecessor had shown apparent representation-space sensitivity in an earlier, uncorrected two-point estimate that likewise did not survive the fuller analysis. The practical reading is that GPT-5.5's OOD advantage (Tables 12,13, Figure 10, which still run on that model) should be understood as tied to target-space extrapolation specifically, not to distributional shift in general.

The k-NN comparisons also carry a practical implication for materials-screening pipelines. Because the LLM's advantage over an independently-tuned k-NN baseline is strongest under compositional shift (LOCO) and least reliable for isolated, sparse test points, a screening workflow gains little by defaulting to the more expensive LLM call for every candidate: k-NN retrieval is a reasonable first-pass filter for candidates close to the training distribution, reserving LLM-based refinement for candidates that represent genuine compositional novelty, where its advantage over k-NN is most consistent. The k-NN-optimal reference-set analysis (Table 15) reinforces this from a different angle: restricting the LLM's context to a small, similarity-selected set of references preserves confidence–error calibration comparable to using the full training set (Random-CV ordered-confidence-ladder ratio of 87.5%), while substantially reducing context length and inference cost. Together, these results point toward a deployable screening strategy — a cheap retrieval step with confidence-gated escalation to LLM reasoning, rather than querying the full model on every candidate — as a direction worth testing explicitly in future work.

The reliability-threshold analysis (Table 14) offers a complementary and more specific form of robustness evidence than the severity slopes alone. Across the full range of severity observed in this study, no model family's fitted error curve ever crosses a trivial baseline that always predicts the training-set mean: 0 of 48 task-level curve pairs cross within the observed range, every model's in-distribution error is already 1.1–3.8× lower than that baseline's, and even at the most severe conditions tested the smallest remaining advantage is 1.11× (LLM-Prop). This is a stronger claim than "error increases with severity" on its own: it shows that, within the shift actually encountered in these datasets, no model tested ever degrades to the point where a materials scientist would do better ignoring its prediction and guessing the training mean instead. Whether this margin persists at severities beyond what these datasets contain is a separate question this analysis cannot answer, and the result should not be read as a guarantee that extends indefinitely.

## Methods

### Data preparation

Data for titanium alloys were compiled from the literature and supplemented with laboratory measurements conducted at XJTU. The dataset comprises 233 entries, each detailing alloy composition, heat treatment processes, and mechanical properties, specifically ultimate tensile strength and elongation. For aluminum alloys, experimental ultimate tensile strength datasets were derived and expanded from Reference[69], resulting in a total of 785 data points including composition, processing, and properties. The high-entropy alloy dataset is related to work in Ref.[70], and comprises 491 FCC based samples characterized by composition, heat treatment and properties. Finally, the steel dataset was adopted from Reference[4]. This collection consists of 677 samples, each encompassing alloy composition, heat treatment types, and mechanical properties (yield strength, ultimate tensile strength, and elongation).

**Model training and hyperparameter optimization**

To establish rigorous predictive baselines, we evaluated three classes of models: conventional machine learning regressors, BERT-embedding-based artificial neural networks, and an LLM-Prop-inspired text-to-property regression model. Two evaluation settings were considered. For the guided–unguided comparison, each property dataset was randomly split into training and testing subsets at an 80:20 ratio using a fixed random seed of 42. The resulting test set was kept unchanged across paired guided and unguided experiments, ensuring that performance differences reflected the effect of guidance rather than variation in data partitioning. For the OOD benchmark, train–test partitions were generated according to the Random CV and OOD target-set generation protocols described in **Target-set generation protocols**. In all cases, model selection and hyperparameter optimization were performed only within the training partition, and the held-out test set was used exclusively for final evaluation.

For traditional machine learning baselines, we considered XGBoost, LightGBM, CatBoost, Random Forest (RF), and Multi-Layer Perceptron (MLP). These models were trained using original numerical descriptors (composition and processing columns) as input features. Hyperparameters were optimized with Optuna[71] using 30 independent trials per model, with 9-fold cross-validation performed within the training partition for model selection. For the tree-based models, the optimized hyperparameters included the number of estimators or boosting iterations, tree depth, learning rate for boosting models, sampling ratios, and regularization-related terms. For MLP, the search space included the hidden-layer architecture, activation function, solver, regularization coefficient $\alpha$, and initial learning rate, with the maximum number of training iterations set to 300.

For BERT family, we employed a BERT-embedding-based ANN framework to integrate compositional and processing information. Chemical compositions and processing-route descriptions were encoded using pre-trained BERT-family models, including SciBERT, MatSciBERT, and SteelBERT. Each encoder produced a 768-dimensional contextual embedding. The composition and processing embeddings were concatenated and passed to a fully connected ANN with nonlinear activation functions, dropout regularization, and a scalar regression output. The ANN was trained using mean squared error loss and the Adam optimizer with weight decay. Early stopping was based on validation loss, and the network architecture and training hyperparameters were optimized using Optuna[71].

We further implemented an LLM-Prop-inspired model following the architecture proposed in LLM-Prop[34]. The model used a T5 encoder initialized from google/t5-v1_1-small[72], followed by pooling and a dropout-regularized linear regression head. Each alloy sample was represented as a textual description containing chemical composition and available processing information. The official LLM-Prop sample checkpoint was not used for formal evaluation; instead, the architecture was fine-tuned separately for each alloy dataset and target property. Target values were z-score normalized using the training data only, and the model was optimized using MAE loss with AdamW and a one-cycle learning-rate schedule. Optuna with 20 trials was used to tune the learning rate, batch size, sequence length, dropout rate, and pooling strategy based on internal validation MAE.

Model performance was evaluated using mean absolute error (MAE), root mean squared error (RMSE), and the coefficient of determination $R^2$: $R^2 = 1 - \frac{\sum_i (y_i - \hat{y}_i)^2}{\sum_i (y_i - \bar{y})^2}$, where $y_i$, $\hat{y}_i$, and $\bar{y}$ denote the observed value, predicted value, and mean observed value, respectively.

**TabPFN-based regression models**

We used TabPFN-2.5-Plus as a pretrained tabular foundation model for alloy property prediction and evaluated two input variants, termed TabPFN-T and TabPFN-N. Both variants shared the same TabPFN-2.5-Plus regression framework but differed in the representation of processing information: TabPFN-T used elemental composition features and text-based processing descriptions, whereas TabPFN-N used the same compositional descriptors together with structured numerical processing variables. Each model was trained separately for each alloy system and target property and evaluated on the corresponding held-out or out-of-distribution test subset using MAE, RMSE, $R^2$ and MAPE.

**Target-set generation protocols**

To evaluate model performance under in-distribution and out-of-distribution conditions, we constructed one in-distribution baseline protocol, Random CV, and six OOD target-set generation protocols: extrapolation, leave-one-cluster-out (LOCO), SparseXsingle, SparseYsingle, SparseXcluster, and SparseYcluster. All target sets were generated independently for each alloy dataset and target property before model-specific feature construction. The target-set generation used the original numeric alloy descriptors, including elemental composition and available numerical heat-treatment or processing descriptors. Target columns, ID-like columns, and metadata columns were excluded from the target-set generation feature space. Missing numeric descriptor values were handled within the preprocessing pipeline, and all X-space descriptors were standardized before clustering, projection, density estimation, and nearest-neighbor search. Based on the k-sensitivity analysis shown in the Fig. S 1, the reported experiments used k = 5 for cluster-based target-set generation.

**Random cross-validation (Random CV)**

Random CV was used as the standard in-distribution evaluation protocol. For each alloy dataset and target property, all valid samples were randomly divided into five approximately equal-sized folds using shuffled k-fold cross-validation with a fixed random seed. In each iteration, one fold, corresponding to approximately 20% of the samples, was held out as the test set, while the remaining four folds, corresponding to approximately 80% of the samples, were used for model training and internal validation. This procedure generated five repeated 80:20 train–test partitions, and each sample was used as a test sample exactly once. The final performance was reported as the average over the five folds. This protocol represents the conventional random-splitting evaluation setting and serves as the in-distribution baseline for assessing model performance under an approximately independent and identically distributed data split.

**Extrapolation**

The extrapolation setting was designed to evaluate model performance under a property-based distribution shift. For each alloy dataset and target property, all samples were sorted according to their target-property values. The samples with the highest target values, corresponding to approximately 20% of the valid data, were selected as the test set, and the remaining samples were used for training. This protocol evaluates the ability of models to extrapolate toward high-property regions that are not represented in the training set.

**Leave-one-cluster-out (LOCO)**

LOCO was used to evaluate model generalization across distinct regions of the alloy input-feature space. For each alloy dataset and target property, samples were represented using the standardized X-space descriptors. The full dataset was clustered into $k= 5$ clusters using k-means in the standardized input-feature space. In each evaluation round, one cluster was held out as the test set, and all remaining clusters were used for model training. Because cluster sizes depend on the distribution of each alloy dataset and target property, the test-set size varied across LOCO folds. Compared with random splitting, LOCO provides a more stringent evaluation by testing models on feature-space-separated groups. However, because all clusters are eventually used for testing, including clusters located in densely populated regions, LOCO may still retain some degree of performance overestimation.

**Single-point targets with the lowest input-feature density (SparseXsingle)**

SparseXsingle was designed to evaluate model performance on isolated samples located in sparse regions of the alloy input-feature space. For each alloy dataset and target property, the standardized input-feature matrix was projected into a two-dimensional space using t-distributed stochastic neighbor embedding (t-SNE). The local density of each sample in this two-dimensional X-space projection was then estimated using Gaussian kernel density estimation. Samples were ranked by increasing X-space density, and a low-density candidate pool was retained, with a maximum candidate pool size of 500.

The candidate samples were then grouped into k = 5 clusters using k-means clustering in the two-dimensional X-space projection. From each candidate cluster, the sample with the lowest X-space density was selected as a representative sparse target point. Thus, SparseXsingle selects isolated low-density input-feature samples while maintaining diversity among the selected points through candidate clustering. The exact selected samples were generated independently for each alloy dataset and target property.

**Single-point targets with the lowest property density (SparseYsingle)**

SparseYsingle identifies sparse target points based on the distribution of target-property values rather than the input-feature distribution. For each alloy dataset and target property, Gaussian kernel density estimation was applied to the one-dimensional target-property distribution to estimate the local Y-space density of each sample. Samples with the lowest property-space densities were retained as sparse-Y candidates, with a maximum candidate pool size of 500.

These sparse-Y candidates were then grouped into k = 5 clusters using k-means clustering in the two-dimensional X-space projection. From each candidate cluster, the sample with the lowest Y-space density was selected as a representative sparse target point. Therefore, SparseYsingle targets rare or underrepresented regions of the property distribution while using X-space clustering to maintain diversity among the selected samples.

**Cluster targets with the lowest input-feature density (SparseXcluster)**

SparseXcluster extends SparseXsingle from isolated target points to local target neighborhoods in sparse input-feature regions. For each alloy dataset and target property, the standardized input-feature matrix was first projected into a two-dimensional space using t-SNE, and local X-space density was estimated using Gaussian kernel density estimation. Samples were ranked by increasing X-space density, and a low-density candidate pool was retained, with a maximum candidate pool size of 500. The candidate samples were then divided into k = 5 groups using k-means clustering in the two-dimensional X-space projection. From each candidate group, the sample with the lowest X-space density was selected as a sparse seed.

To construct local sparse target regions, each seed was expanded in the original standardized input-feature space rather than in the t-SNE space. Specifically, Euclidean distances from the seed to all samples were computed, and the seed itself together with up to five nearest neighbors was considered for inclusion in the test set. If a sample had already been assigned to another seed neighborhood, it was skipped to avoid overlap.

The total number of selected test samples was also constrained by the predefined test-size cap. Therefore, although k = 5 was fixed, the final test-set size and the number of neighbors associated with each seed could vary across alloy datasets and target properties. This protocol evaluates model robustness on local alloy regions that are sparsely represented in the input-feature space.

**Cluster targets with the lowest property density (SparseYcluster)**

SparseYcluster is the property-space counterpart of SparseXcluster. For each alloy dataset and target property, Gaussian kernel density estimation was applied to the one-dimensional target-property distribution, and samples with the lowest Y-space density were retained as sparse-Y candidates. These candidates were then clustered into k = 5 groups using k-means in the two-dimensional X-space projection to maintain feature-space diversity among rare-property candidates. From each group, the sample with the lowest Y-space density was selected as a sparse seed.

Each seed was then expanded into a local test neighborhood using Euclidean distance in the original standardized input-feature space. The seed itself and up to five nearest neighbors were considered for inclusion, while samples already selected by previous seed neighborhoods were excluded to prevent overlap. The total test-set size was further limited by the predefined test-size cap. As a result, SparseYcluster test sets were independently generated for each alloy dataset and target property, and their final sample counts and selected data points were not necessarily identical to those of SparseXcluster. This protocol evaluates model performance on local alloy regions associated with rare or sparsely sampled property values.

**Standardized Wasserstein distance**

To quantify the train–test distributional shift under each OOD split, we computed two standardized Wasserstein distances: $W_1^*(y)$ for the target-property distribution and $W_1^*(x)$ for the input-feature distribution. For a given split, the training and test sets were denoted as

$$D_{\mathrm{tr}}=\{\left(y_i^{\mathrm{tr}},\mathrm{x}_{\mathbf{i}}^{\mathrm{t}}\right)\}_{i=1}^{n},D_{\mathrm{te}}=\{\left(y_j^{\mathrm{te}},\mathrm{x}_{\mathbf{j}}^{\mathrm{t}}\right)\}_{j=1}^{m}, \tag{4}$$

where n and m are the numbers of training and test samples, respectively; y is the target property; and $\mathbf{x}=(x_1,\ldots,x_d)$ is the original input-feature vector, with $d$ denoting the number of features.

For each scalar variable $q\in\{y,x_1,\ldots,x_d\}$, , standardization was performed using only the corresponding training set. The training-set mean and sample standard deviation were computed as

$$\overline{q_{\mathrm{tr}}}=\frac{1}{n}\sum_{i=1}^{n} q_i^{\mathrm{tr}}, s_{q,\mathrm{tr}}=\sqrt{\frac{1}{n\text{-}1}\sum_{i=1}^{n}\left(q_i^{\mathrm{tr}}\text{-}\overline{q_{\mathrm{tr}}}\right)^2}. \tag{5}$$

Both training and test values were then transformed as

$$\widetilde{q_i^{\mathrm{tr}}}=\frac{q_i^{\mathrm{tr}}\text{-}\overline{q_{\mathrm{tr}}}}{s_{q,\mathrm{tr}}}, \widetilde{q_j^{\mathrm{te}}}=\frac{q_j^{\mathrm{te}}\text{-}\overline{q_{\mathrm{tr}}}}{s_{q,\mathrm{tr}}}. \tag{6}$$

The standardized target-property distance was defined as

$$W_1^*(y)=W_1\left(\frac{1}{n}\sum_{i=1}^{n}\delta_{\widetilde{y_i^{\mathrm{tr}}}}, \frac{1}{m}\sum_{j=1}^{m}\delta_{\widetilde{y_j^{\mathrm{te}}}}\right), \tag{7}$$

where $W_1$ denotes the first-order Wasserstein distance, and $\delta_x$ denotes a unit point mass located at sample x. Equivalently, each standardized training sample has weight 1/n, and each standardized test sample has weight 1/m.

For the original input-feature distribution, the standardized vector was $\tilde{\mathbf{x}}=(\tilde{x}_1,\ldots,\tilde{x}_d)$, and the corresponding distance was computed as

$$W_1^*(\mathrm{x})=W_1\left(\frac{1}{n}\sum_{i=1}^{n}\delta_{\widetilde{\mathrm{x}_i^{\mathrm{tr}}}}, \frac{1}{m}\sum_{j=1}^{m}\delta_{\widetilde{\mathrm{x}_j^{\mathrm{te}}}}\right), \tag{8}$$

using the Euclidean ground distance in the standardized input-feature space.

For multi-split OOD strategies, group-level distances were aggregated by test-size-weighted averaging. Because standardization was based exclusively on the training set, both $W_1^*(y)$ and $W_1^*(\mathrm{x})$ are dimensionless and comparable across datasets, target properties, and OOD sampling strategies. Larger values indicate stronger standardized distributional shift and therefore a more severe extrapolation challenge.

The severity ratios reported in Results are $R_x = W_1^*(x)/W_{1,\text{Rand-CV}}^*(x)$ and $R_y = W_1^*(y)/W_{1,\text{Rand-CV}}^*(y)$. The denominator is the same standardized distance computed for the task's random cross-validation splits, so Random-CV itself has a ratio of about 1. Specifically, the raw standardized Wasserstein distances $W_{s,t,\mathrm{RandCV},f}^*(s \in x, y)$are computed independently for each fold $f \in \{1, \ldots, 5\}$, and the denominator $\overline{W}_{s,t,\mathrm{RandCV}}^*$ is calculated as the test-size-weighted average over the five folds, $\overline{W}_{s,t,\mathrm{RandCV}}^* = \sum_{f=1}^{5} n_f\, W_{s,t,\mathrm{RandCV},f}^* / \sum_{f=1}^{5} n_f$, which coincides with the arithmetic mean when folds are equally sized.

t-SNE (perplexity 30, random seed 42) is used to select the seed samples of SparseXsingle and SparseXcluster and to visualize splits (Figure 4); it does not enter the Wasserstein distances, which are computed in the original standardized feature space.

**Density-based control construction**

To determine whether LLM-reported confidence contains information beyond retrieval geometry, we constructed density-based controls using only the retrieved reference samples without additional LLM inference. During the retrieval process, cosine similarities are calculated based on the original standardized numerical feature space between each test sample and its retrieved reference samples. Retrieval density was defined as the average cosine similarity between the target sample and its retrieved k reference samples under uniform weighting:

$$Density_i = \frac{1}{k_i}\sum_{j=1}^{k_i} sim(x_i, x_j) \tag{9}$$

where $k_i$ is the effective number of retrieved neighbors for sample *i*, and $sim(x_i, x_j)$ denotes the cosine similarity between the standardized numerical feature vectors of the target sample and its *j*-th retrieved reference sample.

For the optimal-k setting, density was calculated using the same reference samples provided to the LLM, with the task-specific optimal reference sample number. For the full-reference setting, density was calculated using all available reference samples. To enable a direct comparison with LLM-reported confidence, continuous density values were converted into categorical High, Medium, and Low confidence levels using a rank-based binning strategy. Specifically, within each task–fold–run group, samples were ranked according to density in descending order, and the ranked samples were assigned to High, Medium, and Low categories with the same numbers of samples as the corresponding LLM confidence categories. Therefore, density-based confidence and LLM-reported confidence had identical class proportions, ensuring that the comparison reflected differences in confidence assignment rather than differences caused by unequal bin sizes.

For the additional control Density@optimal-k (full-k bins), density was calculated from the optimal-k retrieved reference samples, while the High, Medium, and Low category sizes were matched to those generated by LLM@full-k. This control isolates whether the confidence–error relationship arises from the density values themselves or from the confidence assignment provided by the LLM.

The ordered confidence-ladder criterion was subsequently applied to both density-based controls and LLM-reported confidence. Confidence-bin MAEs comparisons were performed using the same matched task–protocol units to further evaluate whether LLM-reported confidence provided error-related information beyond retrieval density. Because a low-confidence label is informative only when it correctly anticipates a larger error, the comparison in Table 16 applies an asymmetric criterion: for the high- and medium-confidence bins, the method assigning the lower MAE is taken as the better-discriminating one, whereas for the low-confidence bin, the method assigning the higher MAE is taken as the better-discriminating one.

**LLM configuration and inference settings**

The Gemini series and GPT models were accessed via the OpenRouter API and the Codex API, respectively, while the DeepSeek models were utilized through their official API. To ensure consistency and scalability, we implemented a unified invocation interface using the LiteLLM package. For all evaluations, the temperature was set to T=0, and all other parameters were kept at their default values in the code.

**Data availability**

The datasets used and analysed during the current study can be accessed at https://github.com/Jack-Liu0227/BERT_ML/tree/master/datasets.

**Code availability**

The source code used in this article can be accessed at https://github.com/Jack-Liu0227/BERT_ML and https://github.com/Jack-Liu0227/Fewshot_Guided.

**Acknowledgements**

We are grateful to the HPC Platform, Xi'an Jiaotong University, for support.

**Funding**

T.L acknowledges support from the National Natural Science Foundation of China (No.52350710205), and Z.G. acknowledges support from the National Natural Science Foundation of China (No.52595632 and No.52250191), the Key Research and Development Program of the Ministry of Science and Technology (No.2023YFB4604100).

**Author information**

Authors and Affiliations

State Key Laboratory for Mechanical Behavior of Materials, State Key Laboratory of Porous Metal Materials, School of Materials Science and Engineering,Xi'an Jiatong University, Xi'an, China

YuJie Liu, Dezhen Xue, Jun Sun, Xiangdong Ding, Zhibin Gao & Turab Lookman

AiMaterial Research, Santa Fe, New Mexico, USA

Turab Lookman

Contributions

T.L. and Z.G. conceived the ideas and designed the overall research framework. Y.L. developed the hybrid "Guided Mode" workflow by integrating numerical priors with Large Language Models and implemented the semantic similarity-based retrieval algorithms. Y.L. also performed the benchmarking of general-purpose LLMs against domain-specific models, evaluated the prediction accuracy and uncertainty calibration, and prepared the figures. D.X. provided the Titanium alloy dataset previously assembled from the literature. J.S. and X.D. contributed to discussions. Y.L. and T.L. drafted the manuscript and did the revisions. T.L. supervised the research. All authors read and approved the manuscript.

Corresponding author

Correspondence to Turab Lookman or Zhibin Gao.

**Ethics declarations**

No Competing Interests

Figure 1 Schematic overview of the three-stage hybrid workflow for mechanical property prediction. Stage 1, RAG: the Top-N most similar reference samples are retrieved from the database based on composition and processing information. Stage 2, LLM Inference: the LLM generates the final prediction by comparing the target sample with the retrieved reference samples. "Physics-informed inference" denotes the model's use of physically consistent composition–processing–property relationships drawn from the retrieved reference samples, not an independent mechanistic understanding of materials physics; a representative worked example is given in Supplementary Note.

**Figure 12** Effect of reference-sample selection and reference-set size on UTS prediction. a) Prediction performance for Al and Ti alloy UTS using semantic similarity-based retrieval and random sampling. Panels I, II show MAE for Al and Ti alloys, respectively. Insets show the low ratio range [0.01-0.1]. The green line marks 5% reference ratio. Similarity-based retrieval near-saturated performance with small fraction of high-relevance reference samples, b) Panels I, II show t-SNE projections of retrieved Top 5% and Top 10% reference samples for Ti target samples. Panels III and IV show UTS distributions of reference samples compared with full training set. Green denotes training samples, colored stars target samples and their retrieved references in the embedding space, gray denote full training-set distribution, and blue denote the retrieved reference samples. The blue dashed line is the ground-truth UTS of target sample T#1.

**Figure 13** Effect of reference sample count on LLM prediction performance and comparison with the best-performing k-NN baseline. Panels a–d show the results for Steel-UTS, Matbench-YS, Ti-UTS, and Al-UTS, respectively. Blue circles and connecting lines indicate LLM predictions at different reference sample counts $k$, with $k$ varied from 1 to the corresponding k-NN-optimal $k + 5$; orange diamonds mark the results at a 5% reference budget, and yellow stars indicate the best-performing k-NN baseline. Gray vertical dotted lines mark the corresponding k-NN-optimal $k$. The horizontal axis uses a logarithmic scale, and MAE is reported in MPa. Full denotes the use of all training samples as references. Evaluations were repeated three times for Steel, HEA, Matbench, Ti and Al. The percentages below each panel indicate the relative change in MAE at the 5% reference budget compared with the full-reference setting. Results for all eight tasks are provided in Fig. S27.

**Figure 14** t-SNE plots for standard cross-validation (CV) test set and six out-of-distribution (OOD) test sets generated using different target selection strategies for Ti alloy ultimate tensile strength (UTS) dataset. Each circle represents one of the 233 samples. (a) Five-fold CV with random splitting, where samples are assigned to 5 color-coded test subsets. (b) Leave-one-cluster-out (LOCO) splitting based on clustered sample groups. (c) Extrapolation-based test set distribution. (d) SparseXsingle, where 5 individual test samples are highlighted in distinct colors and the remaining samples are shown in gray. (e) SparseYsingle, where 5 individual test samples are highlighted in distinct colors and the remaining samples are shown in gray. (f) SparseXcluster, where 29 test clusters are highlighted in distinct colors and the remaining samples are shown in gray. (g) SparseYcluster, where 30 test clusters are highlighted in distinct colors and the remaining samples are shown in gray. The figure illustrates how different target-generation methods vary in their coverage of the feature space and in their degree of distributional shift from the full dataset.

**Figure 15** Comparison of the optimized k-NN predictor and GPT-5.5 with matched k-NN references under OOD conditions. For each task–fold configuration, the KNN model was optimized over four representation spaces, neighborhood sizes $k$, and weighting strategies; the configuration with the lowest test-set MAE was selected. Its exact neighbor set was then provided to GPT-5.5 as few-shot references for the same test samples. Each point represents one configuration; boxes and diamonds show the interquartile range and mean, respectively. The Table 10 shows Relative performance of GPT-5.5 and the selected k-NN predictor across OOD protocols using matched KNN references. GPT-5.5 outperformed k-NN in 87 of 120 configurations, while KNN remained superior in 33.

**Figure 16** Model performance comparison for Ti-alloy UTS under Rand-CV and OOD evaluation using full training dataset. a, MAE of six model families, BERT-based, traditional machine-learning models, TabPFN with tabular processing features (TabPFN-N), TabPFN with text-converted processing steps (TabPFN-T), LLM-Prop and GPT-5.5, across Rand-CV and six OOD splits. b, Percentage change in MAE relative to the traditional-model baseline for each split; positive values indicate higher error than Trad., whereas negative values indicate lower error. c, Within-split model ranking based on MAE, where rank 1 denotes best-performing model and rank 5 denotes the worst. d,e, MAE of individual traditional machine-learning models and BERT variants, respectively, illustrating the within-family performance variation across OOD protocols. Lower MAE indicates better predictive performance. The error bars for GPT-5.5 in the figure represent the standard deviation across three runs.

**Figure 17** Representation-space partial-regression analysis across six model families using full training dataset. Panels (a–f) show the partial relationship between representation-space severity and prediction degradation after removing the effects of task identity and target-space severity. According to the Frisch–Waugh–Lovell theorem, the slope of each fitted line is identical to the corresponding coefficient $b_x$ from the full multivariable regression. Red lines denote the fitted partial-regression trends, with the estimated slope and 95% confidence interval reported in each panel. All confidence intervals include zero, indicating no statistically significant independent association between representation-space severity and prediction degradation.

**Figure 18** Target-space partial-regression analysis across six model families using the full training dataset. Panels (a–f) show the partial relationship between target-space severity and prediction degradation after removing the effects of task identity and representation-space severity. Panels (a–f) show the partial relationship between representation-space severity and prediction degradation after removing the effects of task identity and target-space severity. According to the Frisch–Waugh–Lovell theorem, the slope of each fitted line is identical to the corresponding coefficient $b_y$ from the full multivariable regression. Red lines denote the fitted partial-regression trends, with the estimated slope and 95% confidence interval reported in each panel. All slopes are positive and their confidence intervals exclude zero, indicating a statistically significant independent association between target-space severity and prediction degradation.

**Figure 19** Prediction accuracy and error distribution of Gemini-2.5-Flash for titanium alloy property prediction using full training dataset. Scatter plots show predicted versus experimental values for multi-target prediction of (a) UTS and (b) elongation, and single-target prediction of (c) UTS and (d) elongation. Samples are grouped according to the confidence levels reported by the LLM (High: blue; Medium: red). The corresponding MAE and $R^2$ values are shown for each confidence category. Relative error distributions are presented in panels (c) and (d). The bars indicate the number of samples (left axis), whereas the lines represent the corresponding percentages (right axis).

**Figure 20** Confidence–error radar charts across prediction tasks using the k-NN-optimal k reference setting. Each panel shows the mean absolute error (MAE) of LLM predictions under seven evaluation protocols for one alloy property prediction task. The angular axes represent Rand-CV, Extra., LOCO, SX-cls, SX-sgl, SY-cls, and SY-sgl, with their full names and ordering indicated by the central reference radar. The three curves correspond to LLM self-reported confidence levels of high, medium, and low. Radial values represent MAE in the original target units, with MPa for UTS tasks and %

for elongation tasks. Task-specific radial scales are used; therefore, MAE values are directly comparable only within individual panels.

**Figure 21** Effect of OOD severity on LLM-reported confidence and prediction error of GPT-5.5 for the Titanium UTS dataset using full training dataset. a, Proportions of high-, medium- and low-confidence predictions under different dataset splits, with OOD severity ($R_X$). LOCO f1–5 denote five leave-one-cluster-out splits, where f1–f5 correspond to five predefined cluster-level OOD test subsets generated by excluding one cluster at a time from the training data. b, Confidence-level proportions as a function of OOD severity, expressed as $\log_2(R_x)$. c, Mean relative error for each confidence level as a function of OOD severity. Arrows in b and c indicate the sign of the Spearman rank correlation between $log_2(R_x)$ and the plotted quantity, calculated separately for each confidence level using all points with nonzero confidence share, with positive, negative and zero correlations represented by upward-right, downward-right and horizontal-right arrows, respectively. d, Aggregated mean relative error for Rand-CV, all OOD splits and $R_x$ splits. Increasing OOD severity generally reduces the fraction of high-confidence predictions and increases prediction error, while high-confidence predictions remain associated with lower relative-error outcomes.